%% file: arxiv.tex
\documentclass[10pt,twocolumn,letterpaper]{article}

\usepackage[accsupp]{axessibility}

\usepackage[pagenumbers]{cvpr} 

\input{preamble}

\definecolor{cvprblue}{rgb}{0.21,0.49,0.74}
\definecolor{mgreen}{rgb}{0.19,0.80,0.19}
\usepackage[pagebackref,breaklinks,colorlinks,allcolors=cvprblue]{hyperref}
\usepackage{color}
\usepackage{colortbl}
\usepackage{caption}
\usepackage{array}
\usepackage{multirow}
\usepackage{booktabs}
\usepackage{pifont}
\usepackage{amssymb}
\usepackage{amsmath}
\usepackage{bm}
\usepackage{tikz}
\definecolor{Gray}{gray}{0.93}
\usepackage{listings}
\usepackage{algorithm}
\usepackage{algorithmic}
\newlength\savewidth\newcommand\shline{\noalign{\global\savewidth\arrayrulewidth
  \global\arrayrulewidth 1pt}\hline\noalign{\global\arrayrulewidth\savewidth}}

\title{Recover and Match: Open-Vocabulary Multi-Label Recognition through Knowledge-Constrained Optimal Transport}

\author{
\textbf{Hao Tan}\textsuperscript{\rm 1,\rm 2}\footnotemark[1],
\textbf{Zichang Tan}\textsuperscript{\rm 3,\rm 4}\footnotemark[1],
\textbf{Jun Li}\textsuperscript{\rm 2}, 
\textbf{Ajian Liu}\textsuperscript{\rm 2}, 
\textbf{Jun Wan}\textsuperscript{\rm2,\rm5}\footnotemark[2], 
\textbf{Zhen Lei}\textsuperscript{\rm 1,\rm2,\rm5,\rm6}\\
\textsuperscript{\rm 1}SAIS, UCAS
\textsuperscript{\rm 2}MAIS, Institute of Automation, Chinese Academy of Sciences \\
\textsuperscript{\rm 3}SIAT, Chinese Academy of Sciences
\textsuperscript{\rm 4}Sangfor Technologies Inc. \\
\textsuperscript{\rm 5}SAI, UCAS
\textsuperscript{\rm 6}CAIR, HKISI, Chinese Academy of Sciences \\
{\tt\small \{tanhao2023, jun.wan, zhen.lei\}@ia.ac.cn, tanzichang@foxmail.com}
}

\begin{document}
\maketitle

\begin{abstract}
Identifying multiple novel classes in an image, known as open-vocabulary multi-label recognition, is a challenging task in computer vision.
Recent studies explore the transfer of powerful vision-language models such as CLIP.
However, these approaches face two critical challenges:
(1) The local semantics of CLIP are disrupted due to its global pre-training objectives, resulting in unreliable regional predictions.
(2) The matching property between image regions and candidate labels has been neglected, relying instead on naive feature aggregation such as average pooling, which leads to spurious predictions from irrelevant regions.
In this paper, we present RAM (\textbf{R}ecover \textbf{A}nd \textbf{M}atch), a novel framework that effectively addresses the above issues.
To tackle the first problem, we propose Ladder Local Adapter (\textbf{LLA}) to enforce refocusing on local regions, recovering local semantics in a memory-friendly way.
For the second issue, we propose Knowledge-Constrained Optimal Transport (\textbf{KCOT}) to suppress meaningless matching to non-GT labels by formulating the task as an optimal transport problem.
As a result, RAM achieves state-of-the-art performance on various datasets from three distinct domains, and shows great potential to boost the existing methods.
Code: \url{https://github.com/EricTan7/RAM}.
\end{abstract}  

\renewcommand{\thefootnote}{} 
\footnotetext{$^*$ Equal contribution. \quad $^\dagger$ Corresponding author.}

\section{Introduction}
\label{sec:intro}
Open-vocabulary multi-label recognition (OVMLR)~\cite{sun2022dualcoop,xu2022open,he2023open,zhu2024query,liulanguage} is an emerging task in the field of computer vision.
Unlike traditional multi-label or single-label recognition, OVMLR aims to recognize all semantics labels and potential new categories in an image, which offers essential capabilities in vastly changing real-world applications.

\begin{figure*}[t]
    \centering
    \includegraphics[width=0.98\linewidth]{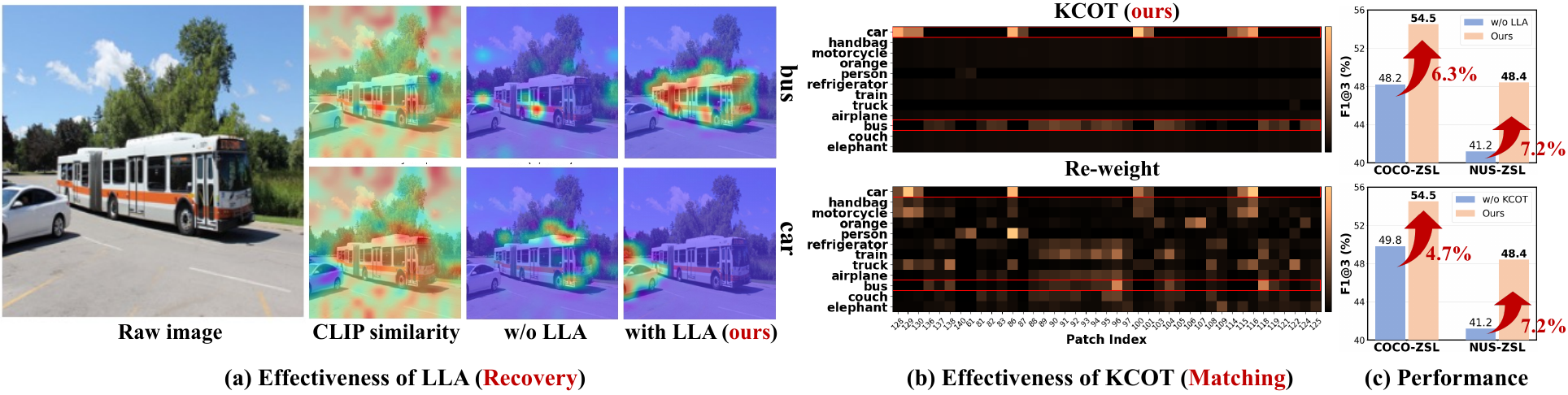}
     \caption{Visualizations and performance comparisons on proposed LLA and KCOT.
     \textbf{(a) Effectiveness of LLA}: CLIP exhibits poor localization capabilities, upon which the optimal transport shows poor results. Our LLA effectively recovers locality, yielding precise matching. 
     \textbf{(b) Effectiveness of KCOT}: we compare the matching between our KCOT and re-weighting in~\cite{sun2022dualcoop} w.r.t. multiple labels. The ground-truth (GT) labels are marked in red boxes. Re-weighting exhibits noisy matching while KCOT precisely focuses on GT labels.
     \textbf{(c) Performance}: both LLA and KCOT bring notable improvements across different datasets.
     }
	\label{fig:intro_all}
\end{figure*}

With the remarkable progress of vision-language models (VLMs)~\cite{radford2021learning, jia2021scaling}, 
recent efforts~\cite{sun2022dualcoop,hu2023dualcoop++,dao2023open,xu2022open,he2023open,zhu2024query,liulanguage} focus on the transfer of CLIP~\cite{radford2021learning}, using different techniques such as PEFT~\cite{zhou2022learning} and knowledge distillation~\cite{hinton2015distilling}.
Specifically, it is common that multiple objects appear in different image regions.
Mismatching between regions and labels leads to spurious predictions from irrelevant areas.
Therefore, \textit{how to construct effective regional features}, and \textit{how to associate these regional features to their corresponding labels} are determinative to the final performance.
However, these two major challenges have been neglected by current approaches~\cite{sun2022dualcoop,hu2023dualcoop++,dao2023open,he2023open,zhu2024query,liulanguage}:
\textbf{(1)} The local semantics in the image encoder are lost~\cite{li2023clip,zhou2023anomalyclip,shao2024explore,dao2023open}.
As evidenced in Figure~\ref{fig:intro_all} (a), the original CLIP exhibits noisy local responses to labels.
Existing methods directly apply these local features, leading to unreliable predictions. 
\textbf{(2)} The matching property between image regions and candidate labels is overlooked.
Instead, they rely on naive region aggregations,
\eg, (a) mean pooling in~\cite{he2023open, zhu2024query}, (b) softmax re-weighting in~\cite{sun2022dualcoop, hu2023dualcoop++} and cross-attention in~\cite{huynh2020shared,liu2021query2label,ma2024text}.
These approaches all fall into \textit{independent re-weighting} since they only consider assignments for each label individually, which exhibits ``extraneous high-response'' phenomenon as shown in Figure~\ref{fig:intro_all} (b), \ie, always emphasizing particular regions for each label including non-GT.
Such phenomenon results in noisy region aggregations.

Based on the above investigations, we propose \textbf{RAM} (\textbf{R}ecover \textbf{A}nd \textbf{M}atch), a conceptually simple yet effective framework, which addresses the above issues through first recovering the local semantics of CLIP and then reformulating the matching as an optimal transport problem.

To address the first issue,
we propose a \textbf{L}adder \textbf{L}ocal \textbf{A}dapter (\textbf{LLA}) to recover local semantics in an efficient way.
Specifically, we enforce the model to refocus on the local context through redesigning the self-attention map and incorporating convolution mechanisms.
Different from most adapters~\cite{sung2022vl,liu2024forgery}, LLA is set as a ladder side network following~\cite{sung2022lst}, which controls the memory cost and preserves the pre-trained knowledge.
As compared in the last two columns in Figure~\ref{fig:intro_all} (a), LLA effectively restores the local information, benefiting the matching process.

To solve the second problem,
we introduce optimal transport (OT) theory~\cite{monge1781histoire, distances2013lightspeed} and propose \textbf{K}nowledge-\textbf{C}onstrained \textbf{O}ptimal \textbf{T}ransport (\textbf{KCOT}), which jointly considers the assignments to all labels, suppressing matching to non-GT labels as shown in Figure~\ref{fig:intro_all} (b).
Specifically, in KCOT, (1) we introduce Label Presence Detection (LPD) to implicitly diminish irrelevant regions by distributing unbalanced marginal constraints, which avoids explicit and rigid selections.
(2) We propose Teacher Knowledge Transfer (TKT) to inject the knowledge of frozen CLIP into the transport process, facilitating accurate matching. 
Notably, KCOT turns out to be an entropic OT problem and can be efficiently solved by Sinkhorn algorithm~\cite{sinkhorn1967concerning}. 
As shown in Figure~\ref{fig:intro_all} (c), both \textit{recovery} (\ie, LLA) and \textit{matching} (\ie, KCOT) bring notable improvements.

To sum up, our main contributions are:
\begin{itemize}
    \item We investigate two limitations in existing OVMLR methods.
    Then we present RAM (\textbf{R}ecover \textbf{A}nd \textbf{M}atch), a simple yet effective framework to address these issues, which achieves substantial gains on various benchmarks
    (\eg, $\geq$+11\% on PA100K and $\geq$+3\% on MS-COCO).
    \item We propose LLA (\textbf{L}adder \textbf{L}ocal \textbf{A}dapter), which recovers local semantics of CLIP in a memory-friendly way, benefiting feature matching in various domains.
    \item We propose KCOT (\textbf{K}nowledge-\textbf{C}onstrained \textbf{O}ptimal \textbf{T}ransport), a reformulation of the region-to-label matching through the lens of optimal transport, which largely enhances generalization on unseen classes.
\end{itemize}

\section{Related Work}
\label{sec:related}
\subsection{Open-Vocabulary Multi-Label Recognition}
Open-Vocabulary Multi-Label Recognition (OVMLR) and Multi-Label Zero-Shot Learning (ML-ZSL) are two similar concepts.
The mutual target is to recognize unseen classes after trained on seen classes.
Traditional ML-ZSL~\cite{zhang2016fast,ben2021semantic,ren2017multiple,rahman2019deep0tag,huynh2020shared,narayan2021discriminative} rely on word embeddings~\cite{mikolov2013distributed, pennington2014glove} to identify unseen labels and focus on the knowledge transfer between seen and unseen labels.
Starting from MIVSE~\cite{ren2017multiple}, regional semantics has been exploited.
Deep0Tag~\cite{rahman2019deep0tag} adopts mix-pooling to aggregate local features.
LESA~\cite{huynh2020shared} and BiAM~\cite{narayan2021discriminative} utilize attention mechanisms for adaptive feature selection.

In the era of large foundation models, OVMLR~\cite{sun2022dualcoop,hu2023dualcoop++,dao2023open,xu2022open,he2023open,zhu2024query,zhang2024diverse,liulanguage} has emerged as a more practical setting.
DualCoOp~\cite{sun2022dualcoop} and DualCoOp++~\cite{hu2023dualcoop++} introduce coupled learnable prompts to transfer the pre-trained knowledge, while the aggregation of features is based on softmax re-weighting.
MKT~\cite{he2023open} filters top-k local features and conducts mean pooling to generate global predictions.
Similarly, QKS~\cite{zhu2024query} explores informative local features while still using mean pooling for aggregation.
CoMC~\cite{liulanguage} focuses on the modality transfer while overlooking the matching property.
These methods fail to jointly consider the affinity between regions and all labels, and ignore the loss of local information in the pre-trained CLIP, leading to unreliable predictions.
Instead, we address the above issues with the proposed KCOT and LLA, respectively.

\begin{figure*}[t]
    \centering
    \includegraphics[width=0.85\linewidth]{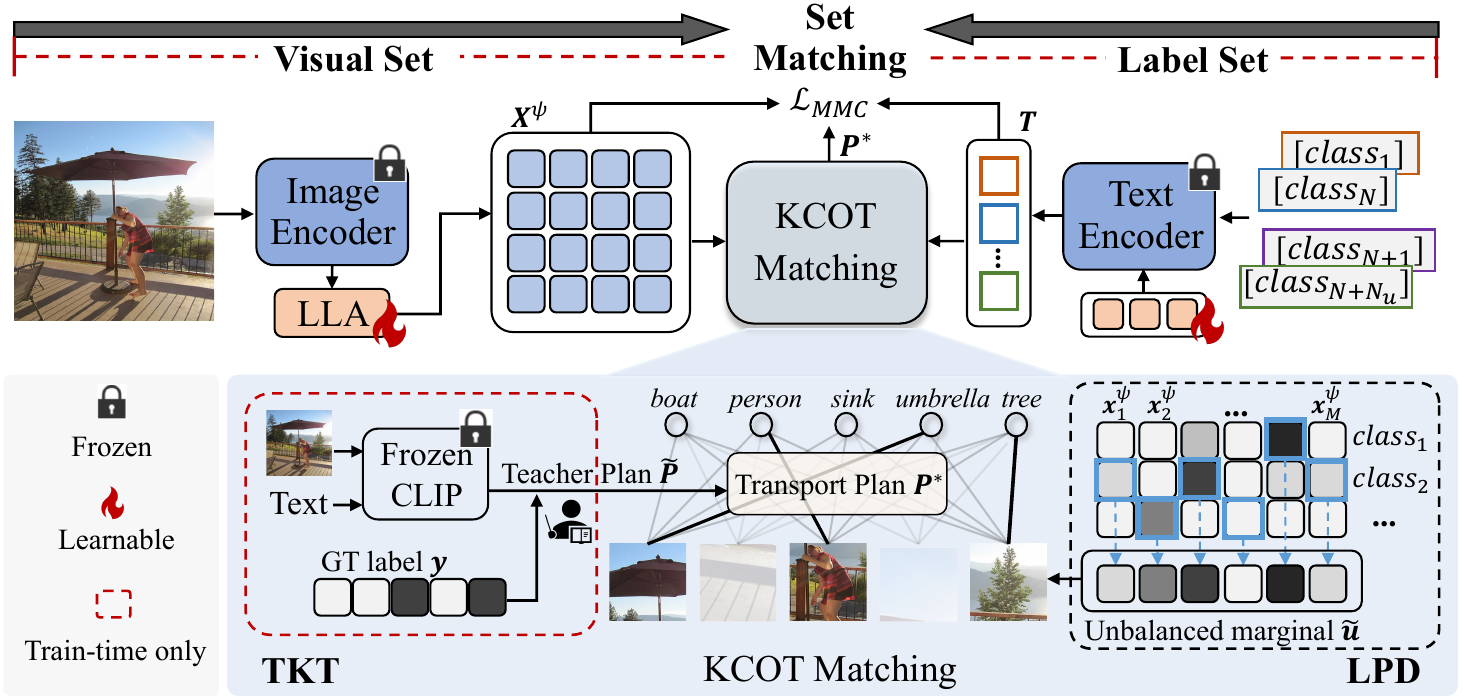}
     \caption{\textbf{Overview of the proposed RAM framework.} 
     LLA is applied to recover the local semantics of the image encoder.
     KCOT is applied between local image features and text features to find a region-to-label matching.
     In KCOT, we propose LPD to distribute unbalanced constraint to visual set, implicitly highlighting foreground areas.
     Moreover, TKT encourages knowledge alignments, and is only performed during training.
     RAM is trained under contrastive objective $\mathcal{L}_{MMC}$.
     The global feature is omitted for clarity.
     }
	\label{fig:main}
\end{figure*}

\subsection{Optimal Transport}
Optimal Transport (OT)~\cite{monge1781histoire} seeks to find the most efficient way to map one probability distribution to another by minimizing a given cost function.
Since the marginal constraints are too strict, various relaxations have been proposed, \eg, entropic optimal transport (EOT)~\cite{distances2013lightspeed} and unbalanced optimal transport (UOT)~\cite{liero2018optimal}, which can be solved iteratively to approximate the optimal solution.
Recent years have seen the increasing applications of these relaxed counterparts in computer vision and machine learning, such as domain adaptation~\cite{alvarez2020geometric} and structural matching~\cite{zhao2021towards}.
Xu et al.~\cite{xu2024temporally} introduce fused Gromov-Wasserstein OT to enhance temporal consistency in action segmentation.
Chen et al.~\cite{chen2022plot} adopt EOT for prompt learning.
PatchCT~\cite{li2023patchct} applies conditional OT for closed-set multi-label recognition, where OT is only adopted for train-time alignments and fails to benefit the inference.
In this paper, we propose a unified OT framework to benefit open-vocabulary multi-label recognition.

\subsection{Vision-Language Models and Efficient Tuning}
Vision-language models (VLMs)~\cite{radford2021learning,jia2021scaling}, pre-trained on large-scale image-text data, have demonstrated remarkable performance across various applications~\cite{luo2022clip4clip, liu2024cfpl}.
Efficient transfer of such pre-trained knowledge to downstream tasks has gained continuous attention.
PEFT (\eg, prompt learning~\cite{zhou2022learning}, adapter tuning~\cite{gao2024clip} and LoRA~\cite{hu2022lora}) has emerged as a powerful technique.
CoOp~\cite{zhou2022learning} and CoCoOp~\cite{zhou2022conditional} introduce learnable textual prompts and yield impressive few-shot performance.
Although the number of extra parameters is limited, these methods are memory-intensive~\cite{tan2024compound}.
CLIP-adapter~\cite{gao2024clip} instead applies MLP for post-refinement, while their zero-shot abilities are hindered.
Distinct from them, we propose LLA, which enhances zero-shot multi-label performance with efficient memory usage.

\section{Preliminaries}

\noindent \textbf{Notations.}
Suppose a mini-batch input images $\{\bm{I}_b\}_{b=1}^{B}$ are labeled with $N$ seen classes. $\bm{y}^{(b)}\in \mathbb{R}^{N}$ denotes the multi-hot label vector, where $\bm{y}_i^{(b)}=1$ means the image $\bm{I}_b$ contains the $i^{th}$ label and vice versa.
In the following we drop the batch notations for simplicity.
The image encoder and text encoder pre-trained from CLIP are denoted as $\mathcal{F}^{\mathcal{V}}(\cdot)$ and $\mathcal{F}^{\mathcal{T}}(\cdot)$, respectively.
The visual features are extracted as $[\bm{x}^g, \bm{X}]=\mathcal{F}^{\mathcal{V}}(\bm{I})$, where $\bm{x}^g\in \mathbb{R}^{d}$ denotes the global visual feature and $\bm{X}=\{\bm{x}_k|\bm{x}_k\in \mathbb{R}^d\}_{k=1}^{M}$ denotes $M$ regional visual features.
Given $N$ seen classes in the training set, our goal is to generalize to arbitrary $N+N_u$ classes, where $N_u>0$ denotes the number of unseen classes.

\noindent \textbf{Optimal Transport.}
Optimal transport (OT)~\cite{monge1781histoire} aims to find the most efficient plan for transferring mass between two distributions.
In this paper, we only consider the discrete situation which is closely related to our framework.
Solving the original OT problem is time-consuming.
By introducing entropy relaxation, the solution can be iteratively approximated by the Sinkhorn algorithm~\cite{sinkhorn1967concerning}.
The entropic OT problem~\cite{distances2013lightspeed} is formulated as:
\begin{equation}
\begin{aligned}
    &\mathop{\min}_{\bm{P}} \sum_{k=1}^{M} \sum_{i=1}^{N} \bm{P}_{ki} \bm{C}_{ki} - \lambda H(\bm{P}), \\
    &\ \text{s.t.}\ \ \bm{P} \bm{1}^N=\bm{u},\ \bm{P}^\mathsf{T} \bm{1}^M=\bm{v},
\end{aligned}
\label{eq:eot}
\end{equation}
where $\bm{C}$ is the cost matrix measuring the distance between two sets of features, such as $\bm{C}_{ki}=1-\text{cos}(\bm{f}_k, \bm{g}_i)$.
$\bm{P}$ denotes the transport plan which is learned under the marginal constraints.
$H(\bm{P})=-\sum_{ki}\bm{P}_{ki}\log(\bm{P}_{ki})$ is the entropy and $\lambda$ denotes the regularization parameter.
$\bm{u}=\{\bm{u}_k\}_{k=1}^{M}$ and $\bm{v}=\{\bm{v}_i\}_{i=1}^{N}$ depict the weights of features in the two sets, which are $M$ and $N$-dimensional simplex, respectively.
Typically, $\bm{u}$ and $\bm{v}$ are set as uniform weights for a balanced transport, \ie, $\bm{u}=\{\bm{u}_k|\bm{u}_k=\frac{1}{M}\}_{k=1}^{M}$ and $\bm{v}=\{\bm{v}_i|\bm{v}_i=\frac{1}{N}\}_{i=1}^{N}$.

\section{Method}
\label{sec:method}
As shown in Figure~\ref{fig:main}, the framework of RAM is conceptually simple.
The label set is constructed with learnable prompts (Sec.~\ref{sec:label}).
The visual set is constructed with LLA (Sec.~\ref{sec:visual}).
KCOT is applied to find a precise region-to-label matching (Sec.~\ref{sec:matching}).
Then, the similarities between visual and label sets are weighted by the matching result, which serve as predictions in training and inference (Sec.~\ref{sec:loss}).

\subsection{Deep Label Prompting}
\label{sec:label}
In this part, we aim to construct the label set.
Manual text prompts are hard to generalize to particular domain without delicate designs.
Alternatively, CoOp~\cite{zhou2022learning} introduces learnable prompts to release human efforts.
We inherit their advantages while re-designing the location of prompts to reduce memory consumption during training.
Suppose there are $L_t$ transformer layers in the text encoder, we introduce $r$ learnable prompt tokens $\bm{Q}_{l}=\{\bm{q}_l^i|\bm{q}_l^i\in \mathbb{R}^{d}\}_{i=1}^{r}$ only in the last few layers.
For example, for the $l^{th}$ layer, the prompt tokens are prepended to the word embeddings:
\begin{equation}
    \bm{T}_{l} = \mathcal{F}_{l}^{\mathcal{T}}([\bm{Q}_{l}, \bm{E}_{l}]),\quad l=l_s,...,L_t,
\end{equation}
where $\mathcal{F}_{l}^{\mathcal{T}}(\cdot)$ is the $l^{th}$ text layer and $[\cdot]$ denotes the concatenation.
$l_s$ is the starting layer and $\bm{E}_l=[\bm{e}_{l}^1,\bm{e}_{l}^2,...,\bm{e}_{l}^h]$ denotes the word embeddings of original input.
The features from the last layer are collected to construct label set: $\bm{T}=\{\bm{t}_i|\bm{t}_i\in \mathbb{R}^{d}\}_{i=1}^{N}$.
Such design saves memory usage by shortening the gradient flow while keeping limited learnable parameters, which enables efficient transfer to OVMLR.

\subsection{Ladder Local Adapter}
\label{sec:visual}
\begin{figure}[t]
    \centering
    \includegraphics[width=0.85\linewidth]{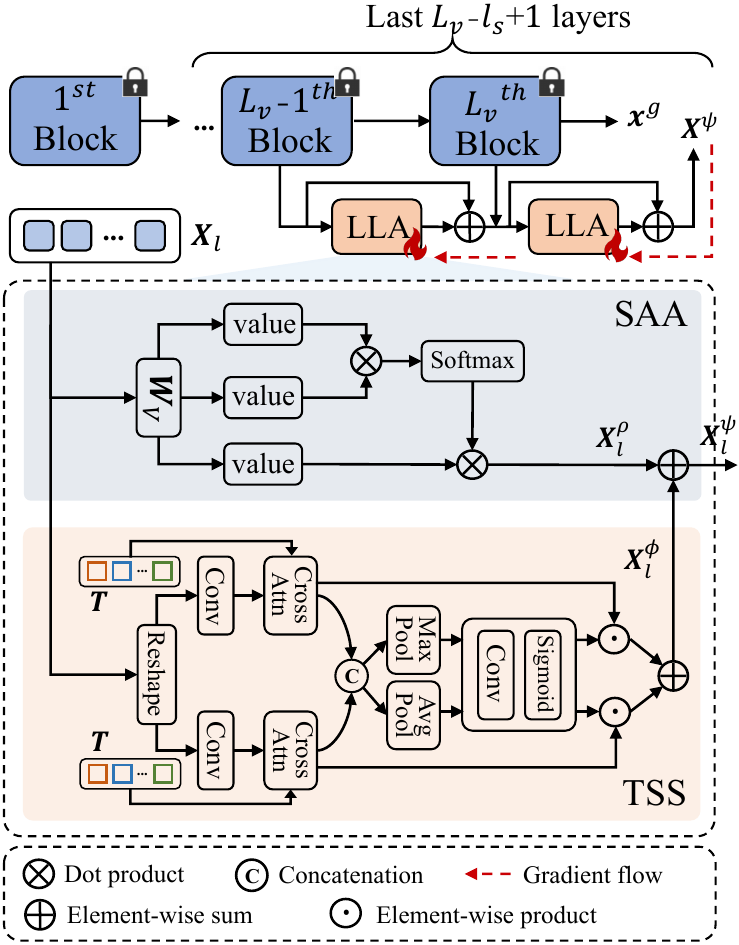}
     \caption{\textbf{The proposed Ladder Local Adapter (LLA).} 
     There are no connections \textit{back} to the original image encoder (\ie, ladder side structure), where the gradient is only propagated within LLA, enabling efficient transfer.
     The outputs from SAA and TSS are averaged.
     LLA is only applied in the last few layers.
     }
	\label{fig:tla}
    \vspace{-0.1cm}
\end{figure}

In this part, we aim to construct the visual set.
Previous works~\cite{li2023clip,zhou2023anomalyclip,shao2024explore} have shown that regional information in CLIP has been diminished.
To facilitate precise matching, we first propose Ladder Local Adapter (LLA) to recover discriminative local semantics.
As shown in Figure~\ref{fig:tla}, LLA consists of two parts, \ie, self-aware attention and text-guided spatial selection.

\begin{figure}[t]
    \centering
    \includegraphics[width=1.0\linewidth]{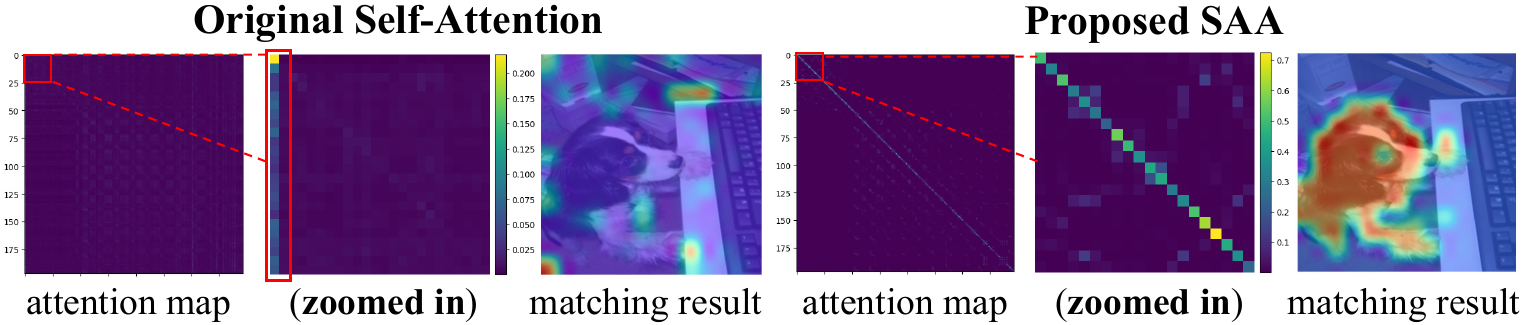}
    \caption{The comparison of attention maps between original self-attention (left) and our SAA (right). Original attention is overwhelmed by dominant patch, while SAA produces diagonal-style attention maps (see Appendix~\ref{app:dis_saa} for more discussions).}
	\label{fig:saa}
    \vspace{-0.4cm}
\end{figure}

\noindent \textbf{Self-aware attention (SAA).}
As shown in Figure~\ref{fig:tla}, 
Different from standard self-attention, SAA computes attention map from parameter \textit{value} itself.
As presented in Figure~\ref{fig:saa},
this makes sure the attention score on itself is the highest, and nearby regions are highlighted,
which enforces the attention to refocus on local regions.
Suppose there are $L_v$ transformer layers in the visual encoder, SAA for the $l^{th}$ layer is denoted as:
\begin{equation}
    \bm{X}_{l}^{\rho}=\mathcal{F}_{\text{SAA}}({\bm{X}_{l}}),\quad l = l_s, ..., L_v.
\end{equation}
Note that SAA does not involve any extra parameters, serving as a rapid and efficient recovery.

\noindent \textbf{Text-guided spatial selection (TSS).}
In TSS, the original features $\bm{X}_{l}$ are first reshaped into feature maps and go through two convolutions.
Then, we query the existence of label semantics in each spatial region by performing a parameter-free cross-attention between spatial features and text features $\bm{T}$.
Conditioned on the semantic-aware spatial features, we generate spatial masks by applying channel-based maximum and average pooling.
The features are then weighted by the masks.
The convolutions in TSS aggregate spatial information with different receptive fields and the masking process helps filter prominent regions that have high response to text features.
We avoid the cumbersome notations in TSS and summarize the above process as:
\begin{equation}
    \bm{X}_{l}^{\phi}=\mathcal{F}_{\text{TSS}}({\bm{X}_{l}}, \bm{T}),\quad l = l_s, ..., L_v. 
\end{equation}

To enable efficient transfer, we parallel the SAA and TSS in a ladder structure, as shown in Figure~\ref{fig:tla}.
Specifically, there are no connections back to the original backbone.
Such design offers two benefits:
\textbf{(1)} \textit{The pre-trained knowledge is largely preserved} since there are no disruptions in the original forward path.
\textbf{(2)} \textit{The memory usage is greatly reduced} as the gradients only propagate on the last few layers of TSS.
The final adapted regional features are the average of SAA and TSS with residual connection:
\begin{equation}
    \bm{X}^{\psi}_{l} = \bm{X}^{\psi}_{l-1} + \frac{1}{2}(\bm{X}^{\rho}_{l}+\bm{X}^{\phi}_{l}),\quad l = l_s, ..., L_v. 
\end{equation}
We collect the adapted local features from the last layer to construct the visual set $\bm{X}^{\psi}=\{\bm{x}^{\psi}_{k}|\bm{x}^{\psi}_{k}\in \mathbb{R}^{d}\}_{k=1}^{M}$, which contains $M$ discriminative regional features.

\subsection{Knowledge-Constrained Optimal Transport}
\label{sec:matching}
\begin{figure}[t]
    \centering
    \includegraphics[width=1.0\linewidth]{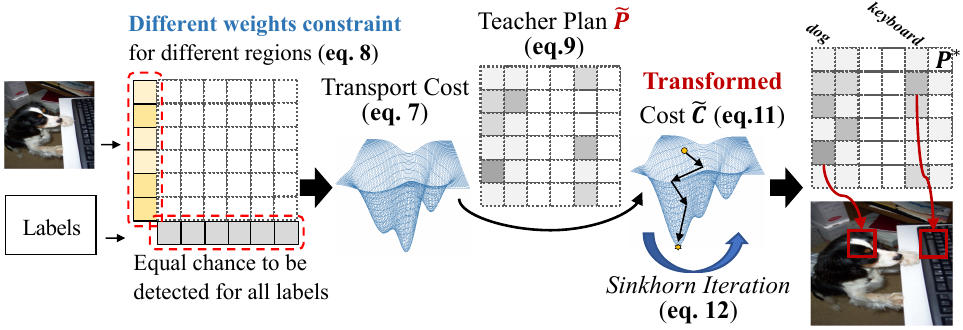}
    \caption{The illustration of KCOT process. LPD delivers weights constraint on image regions. TKT simply modifies the cost.}
	\label{fig:kcot}
    \vspace{-0.4cm}
\end{figure}

In this part, we aim to find a precise matching between visual and label set.
A straightforward way is to find a one-to-one matching, while such sparse matching leads to unsatisfactory performance (Table~\ref{tab:abl_match}).
In this paper, we formulate the task as an optimal transport problem, as shown in Figure~\ref{fig:kcot}.
This offers two benefits:
(1) the affinity between regions and all labels are jointly compared, which suppresses matching to non-GTs.
(2) By introducing unbalanced marginal and teacher assignment to guide the transport, we derive the knowledge-constrained optimal transport (KCOT) problem, which enhances generalization to unseen classes.
Specifically, the KCOT is formulated as:
\begin{equation}
\begin{aligned}
    &\mathop{\min}_{\bm{P}} \underbrace{\sum_{k=1}^{M} \sum_{i=1}^{N} \bm{P}_{ki} \bm{C}_{ki}}_{\text{transport cost}}  - \underbrace{\lambda_1 H(\bm{P})}_{\text{entropy term}} + \underbrace{\lambda_2 D_{KL}(\bm{P}||\bm{\widetilde{P}})}_{\text{knowledge injection}}, \\
    &\ \text{s.t.}\ \ \bm{P} \bm{1}^N=\bm{\tilde{u}},\ \bm{P}^\mathsf{T} \bm{1}^M=\bm{v},
\end{aligned}
\label{eq:kcot}
\end{equation}
where $\bm{\widetilde{P}}$ is a teacher plan. $D_{KL}(\cdot)$ denotes the Kullback-Leibler (KL) divergence.
$\bm{\tilde{u}}$ is unbalanced marginal for visual set and $\bm{v}$ is uniform marginal for label set.

\noindent \textbf{Transport Cost.}
Given the visual set $\bm{X}^{\psi}$ and the label set $\bm{T}$, the transport cost between the $k^{th}$ image region and the $i^{th}$ label is defined as:
\begin{equation}
    \bm{C}_{ki} = 1 - \frac{\text{exp}(\text{cos}(\bm{x}_k^{\psi}, \bm{t}_i)/\tau)}{\sum_{j=1}^{N}\text{exp}(\text{cos}(\bm{x}_k^{\psi}, \bm{t}_j)/\tau))},
\end{equation}
where $\text{cos}(\cdot)$ denotes the cosine similarity.
$\tau$ is the pre-trained temperature.
In our studies, the normalized and reversed similarities serve as an appropriate cost for KCOT.

\noindent \textbf{Label Presence Detection (LPD).}
The uniform hypothesis of the marginals (\ie, $\bm{u}=\{\bm{u}_k|\bm{u}_k=\frac{1}{M}\}_{k=1}^{M}$) forces all image regions to have the same importance.
However, this is not applicable to OVMLR since there are many irrelevant regions that are far from the labels.
In KCOT, we first assess the existence of labels in each region.
Those regions with low existence scores are \textit{implicitly} diminished by distributing unbalanced weights in the marginal constraints:
\begin{equation}
    \bm{\tilde{u}}_k = \frac{\text{exp}(\mathop{\max}_{i}((\text{cos}(\bm{x}_k^{\psi}, \bm{t}_i)/\tau))}{\sum_{m=1}^{M}\text{exp}(\mathop{\max}_{i}(\text{cos}(\bm{x}_m^{\psi}, \bm{t}_i)/\tau)))},
\end{equation}
where $\mathop{\max}(\cdot)$ retrieves the most relevant class for current region to detect the presence of any label.
These similarities are compared along all regions to obtain the weights.
Different from top-k selections~\cite{li2023patchct,lafon2024gallop}, LPD \textit{prioritizes important regions by marginal constraint in OT}, avoiding rigid and explicit selection.
Marginal $\bm{v}$ remains uniform to guarantee any unseen class has an equal chance to be identified.

\noindent \textbf{Teacher Knowledge Transfer (TKT).}
It is non-trivial to find a perfect teacher plan $\bm{\widetilde{P}}$ in eq.~\ref{eq:kcot}.
Ideally, the transport weights between ground-truth labels and their relevant regions should be large, which for negative labels should be close to zero.
To this end, we suppose frozen CLIP is a good supervisor for its well-aligned feature spaces and versatile knowledge.
To approach an ideal $\bm{\widetilde{P}}$, we further take the ground-truth labels to mask the teacher assignment:
\begin{equation}
\begin{aligned}
    &\bm{P}^{\prime}_{ki} = \frac{\text{exp}(\text{cos}(\bm{x}^{\prime}_k, \bm{t}^{\prime}_i)/\tau)}{\sum_{j=1}^{N}\text{exp}(\text{cos}(\bm{x}^{\prime}_k, \bm{t}^{\prime}_j)/\tau))}, \\
    &\bm{\widetilde{P}}_{ki} =
        \begin{cases}
            \bm{P}^{\prime}_{ki}, &\bm{y}_i=1, \\
            \mathop{\min}_{k,i}(\bm{P}^{\prime}_{ki}), &\bm{y}_i=0,
        \end{cases}
\end{aligned}
\end{equation}
where $\bm{x}^{\prime}$ and $\bm{t}^{\prime}$ are the visual and text features extracted by the frozen CLIP, and the SAA is applied to get $\bm{x}^{\prime}$.
Masked by label $\bm{y}$, the assignment to absent labels remains minimal, which imitates an ideal assignment.
Note that TKT is only applied during training to facilitate cross-modal alignments.

\begin{table*}[t]
    \centering
    \scalebox{0.94}{
        \small
        \begin{tabular}{p{62pt}<{\raggedright}|p{18pt}<{\centering}p{18pt}<{\centering}p{18pt}<{\centering}p{18pt}<{\centering}p{18pt}<{\centering}p{18pt}<{\centering}p{20pt}<{\centering}|p{18pt}<{\centering}p{18pt}<{\centering}p{18pt}<{\centering}p{18pt}<{\centering}p{18pt}<{\centering}p{18pt}<{\centering}p{20pt}<{\centering}}
        \multirow{2}{*}{\hspace{-5pt}Method} & \multicolumn{7}{c}{Zero-shot Learning (ZSL)} & \multicolumn{7}{c}{Generalized Zero-shot Learning (GZSL)} \\
        \cline{2-15}
        & \rule{0pt}{8pt}P@3 & R@3 & F1@3 & P@5 & R@5 & F1@5 & mAP & P@3 & R@3 & F1@3 & P@5 & R@5 & F1@5 & mAP\\
        \shline
        \hspace{-5pt}\rule{0pt}{7pt}LESA$_{\text{M10}}$~\cite{huynh2020shared} & 25.7 & 41.1 & 31.6 & 19.7 & 52.5 & 28.7 & 19.4 & 23.6 & 10.4 & 14.4 & 19.8 & 14.6 & 16.8 & 5.6\\
        \hspace{-5pt}BiAM~\cite{narayan2021discriminative} & - & - & 33.1 & - & - & 30.7 & 26.3 & - & - & 16.1 & - & - & 19.0 & 9.3\\
        \hspace{-5pt}SDL$_{\text{M7}}$~\cite{ben2021semantic} & 24.2 & 41.3 & 30.5 & 18.8 & 53.4 & 27.8 & 25.9 & 27.7 & 13.9 & 18.5 & 23.0 & 19.3 & 21.0 & 12.1\\
        \hspace{-5pt}CLIP~\cite{radford2021learning} & 33.5 & 41.6 & 37.1 & 25.1 & 51.8 & 33.8 & 39.4 & 27.2 & 11.9 & 16.5 & 22.2 & 16.2 & 18.7 & 16.8 \\
        \hspace{-5pt}DualCoOp~\cite{sun2022dualcoop} & 37.3 & 46.2 & 41.3 & 28.7 & 59.3 & 38.7 & 43.6 & 31.9 & 13.9 & 19.4 & 26.2 & 19.1 & 22.1 & 12.0\\
        \hspace{-5pt}DualCoOp++~\cite{hu2023dualcoop++} & \underline{42.4} & 52.5 & \underline{46.9} & \underline{31.2} & 64.5 & \underline{42.1} & 47.1 & 34.7 & 15.2 & 21.1 & 29.2 & 21.3 & 24.7 & 15.1\\
        \hspace{-5pt}OVML~\cite{dao2023open} & 36.3 & 44.7 & 40.1 & 27.9 & 57.2 & 37.5 & 42.6 & 32.9 & 12.6 & 18.3 & 28.1 & 18.0 & 22.0 & 14.3\\
        \hspace{-5pt}MKT~\cite{he2023open} & 27.7 & 44.3 & 34.1 & 21.4 & 57.0 & 31.1 & 37.6 & \underline{35.9} & \underline{15.8} & \underline{22.0} & \underline{29.9} & \underline{22.0} & \underline{25.4} & \underline{18.3} \\
        \hspace{-5pt}TGF~\cite{ma2023transferable} & 29.0 & 46.3 & 35.6 & 21.4 & 56.9 & 31.1 & 31.1 & 33.9 & 14.9 & 20.7 & 29.1 & 21.4 & 24.6 & 15.8 \\
        \hspace{-5pt}CoMC~\cite{liulanguage} & 33.5 & \underline{53.5} & 41.2 & 24.8 & \underline{66.1} & 36.1 & \underline{48.2} & - & - & - & - & - & - & - \\
        \rowcolor{Gray}\hspace{-5pt}RAM (\textbf{ours}) & \textbf{43.7} & \textbf{54.2} & \textbf{48.4} & \textbf{33.8} & \textbf{69.9} & \textbf{45.6} & \textbf{53.2} & \textbf{38.7} & \textbf{16.9} & \textbf{23.5} & \textbf{32.7} & \textbf{23.9} & \textbf{27.6} & \textbf{20.7} \\ 
        \rowcolor{Gray}\hspace{-5pt}\textbf{$\Delta$ Prev. Best} & \textcolor{red}{\textbf{+1.3}} & \textcolor{red}{\textbf{+0.7}} & \textcolor{red}{\textbf{+1.5}} & \textcolor{red}{\textbf{+2.6}} & \textcolor{red}{\textbf{+3.8}} & \textcolor{red}{\textbf{+3.5}} & \textcolor{red}{\textbf{+5.0}} & \textcolor{red}{\textbf{+2.8}} & \textcolor{red}{\textbf{+1.1}} & \textcolor{red}{\textbf{+1.5}} & \textcolor{red}{\textbf{+2.8}} & \textcolor{red}{\textbf{+1.9}} & \textcolor{red}{\textbf{+2.2}} & \textcolor{red}{\textbf{+2.4}} \\
        \end{tabular}
    }
    \vspace{-0.25cm}
    \caption{\textbf{Results (\%)} on the open-vocabulary NUS-WIDE. ZSL refers to testing on unseen classes. GZSL refers to unified testing on both seen and unseen classes. The best scores are \textbf{bolded}. The second best scores are \underline{underlined}.}
    \label{tab:main_nus}
    \vspace{-0.5cm}
\end{table*}

\noindent \textbf{Problem Solver.}
Conditioned on the definitions of KL-divergence and entropy, eq.~\ref{eq:kcot} can be expressed as follows (see Appendix~\ref{app:deriv_kcot} for derivation):
\begin{equation}
\begin{aligned}
    &\mathop{\min}_{\bm{P}} \sum_{k=1}^{M} \sum_{i=1}^{N} \bm{P}_{ki} \bm{\widetilde{C}}_{ki} -\tilde{\lambda} H(\bm{P}), \\
    &\ \text{s.t.}\ \ \bm{P} \bm{1}^N=\tilde{\bm{u}},\ \bm{P}^\mathsf{T} \bm{1}^M=\bm{v},
\end{aligned}
\label{eq:kcot_final}
\end{equation}
where $\tilde{\lambda}=\lambda_1+\lambda_2$ and the transformed cost is:
\begin{equation}
    \bm{\widetilde{C}}_{ki} = \bm{C}_{ki} - \lambda_2\log\bm{\widetilde{P}}_{ki}.
\end{equation}
Eq.~\ref{eq:kcot_final} is equivalent to an entropic OT problem, which can be solved by Sinkhorn~\cite{sinkhorn1967concerning} within several iterations:
\begin{equation}
    \bm{P}^{*} = \text{diag}(\bm{a}^t)\text{exp}(-\bm{\widetilde{C}}/\tilde{\lambda})\text{diag}(\bm{b}^t),
\end{equation}
where $t$ denotes the iteration.
$\bm{a}^t=\bm{\tilde{u}}/((\text{exp}(-\bm{\widetilde{C}}/\tilde{\lambda})\bm{b}^{t-1})$ and
$\bm{b}^t=\bm{v}/((\text{exp}(-\bm{\widetilde{C}}/\tilde{\lambda})^{\mathsf{T}}\bm{a}^{t})$, with initialization on $\bm{b}^0=\bm{1}$.
$\bm{P}^{*}$ is the optimal transport plan with $\bm{P}^{*}_{ki}$ denoting the matching weight of the $k^{th}$ image region to the $i^{th}$ label.

\subsection{Training and Inference}
\label{sec:loss}
In line with the set matching framework, we propose Multi-Matching Contrastive (MMC) loss.
For a mini-batch images $\{\bm{I}_b\}_{b=1}^{B}$, the predicted logit of the $i^{th}$ label in $\bm{I}_b$ is $s(b, i)$.
Inspired by~\cite{khosla2020supervised, oord2018representation}, MMC loss is defined as:
\begin{small}
\begin{equation}
    \mathcal{L}_{\text{MMC}}\!=\!-\frac{1}{|\mathcal{P}_B|}\!\sum_{(b,i)\in \mathcal{P}_B}\!\!\!\!\log \frac{\text{exp}(s(b,i)/\tau')}{\sum_{b^{'}\!=1}^{B}\!\sum_{j=1}^{N}\!\text{exp}(s(b^{'}\!,j)/\tau')},
    \label{eq:mmc}
\end{equation}
\end{small}
where $\mathcal{P}_B= \{(b,i)|\bm{y}_i^{(b)}=1 \}$ is the set of indices of all positive pairs in the mini-batch and $|\mathcal{P}_B|$ is its cardinality.
$\tau'$ is a learnable temperature parameter.

During training, we consider the global feature as a special region that adaptively aggregates information of the whole image.
The prediction from global feature is its similarity with text features.
The predictions from local regions are aggregated based on the matching result $\bm{P}^{*}$:
\begin{equation}
\begin{aligned}
    &s^R(b, i) = \sum_{k=1}^{M}\bm{P}^{*(b)}_{ki}\, \text{cos}(\bm{x}_{k}^{\psi (b)}, \bm{t}_i), \\
    &s^G(b, i) = \text{cos}(\bm{x}^{g (b)}, \bm{t}_i),
\end{aligned}
\end{equation}
where $s^R(b, i)$ and $s^G(b, i)$ are predictions from local and global feature, respectively.
Final training objective is the averaged loss of $s^R(b, i)$ and $s^G(b, i)$ derived from eq.~\ref{eq:mmc}:
\begin{equation}
    \mathcal{L} = \mathcal{L}_{\text{MMC}}^{R} + \mathcal{L}_{\text{MMC}}^{G}.
\end{equation}

During inference, only LPD is applied in KCOT.
$s^R(b, i)$ and $s^G(b, i)$ are normalized and averaged to get final score.

\section{Experiments}
\label{sec:exp}

\subsection{Datasets and Metrics}
\textbf{Datasets.}
Following~\cite{bansal2018zero, ben2021semantic, huynh2020shared}, we conduct experiments on MS-COCO~\cite{lin2014microsoft} and NUS-WIDE~\cite{chua2009nus}.
We take both zero-shot learning (ZSL) and generalized zero-shot learning (GZSL) tasks.
ZSL only tests on unseen classes and GZSL tests on both seen and unseen classes.
\textbf{MS-COCO} is split into 48 seen classes and 17 unseen classes.
\textbf{NUS-WIDE} is split into 925 seen classes and 81 unseen classes.

\noindent We also evaluate our method on other domains including pedestrian attribute recognition (PAR) and remote sensing (RS) image classification.
We take the most popular datasets, \ie, \textbf{RAPv1}~\cite{li2018richly} and \textbf{PA100K}~\cite{deng2014pedestrian} for PAR, \textbf{MultiScene}~\cite{hua2021multiscene} and \textbf{MLRSNet}~\cite{qi2020mlrsnet} for RS.
We take GZSL task, where 30\% classes are sampled as unseen labels, deriving OV benchmarks.
More details please refer to Appendix~\ref{app:details}.

\noindent \textbf{Evaluation Metrics.}
Following~\cite{ben2021semantic, huynh2020shared}, we report top-3 recall (R@3), precision (P@3), F1 score (F1@3) and mAP on COCO.
On NUS-WIDE, we further report top-5 recall (R@5), precision (P@5) and F1 score (F1@5).
On other datasets, we report F1@3, F1@5 and mAP.

\subsection{Implementation Details}
We take ViT-B/16 model pre-trained from CLIP~\cite{radford2021learning}.
Both the image and text encoders remain frozen during training.
For a fair comparison, the input images are resized into 224 $\times$ 224.
$\lambda_1$ and $\lambda_2$ are set as 0.1 and 0.05, respectively.
The maximum iterations in KCOT is set to 100.
LLA and textual prompts are only applied in the last three layers with $L_v=L_t=12$ and $l_s=10$.
Prompt length $r$ is set to 4.
The model is trained for 6 epochs with a batch size of 32.
For comprehensive comparisons, we reproduced previous methods~\cite{narayan2021discriminative, sun2022dualcoop, ridnik2023ml, he2023open} and specialized model RemoteCLIP~\cite{liu2024remoteclip} on domain benchmarks.

\subsection{Main Results}
\noindent \textbf{Comparisons on natural images.}
As shown in Table~\ref{tab:main_nus} and Table~\ref{tab:main_coco}, our method achieves SOTA performance across all metrics on both NUS-WIDE and MS-COCO.
Specifically, RAM achieves significant improvements on both ZSL and GZSL, surpassing previous SOTA by 5.0\% mAP on NUS-ZSL and 3.3\% F1@3 on COCO-GZSL.
Compared to recent methods~\cite{he2023open, sun2022dualcoop, hu2023dualcoop++, liulanguage} that overlook the matching property, RAM achieves notable gains, demonstrating the effectiveness of our set matching framework.

\noindent \textbf{Comparisons on other domains.}
As shown in Table~\ref{tab:main_par} and Table~\ref{tab:main_rs}, RAM achieves tremendous improvements, \eg, surpassing previous SOTA by 9.6\% and 15.3\% F1@3 on MLRSNet-OV and PA100K-OV, respectively.
Notably, the performance of previous methods and zero-shot CLIP is extremely limited due to domain gaps, \eg, MKT~\cite{he2023open} which is built on CLIP, achieves less than 20\% F1@3 on PAR benchmarks.
In contrast, RAM achieves robust performance, highlighting its superiority and versatility.

\subsection{Ablation Studies}
We conduct ablation studies on both ZSL and GZSL tasks of NUS-WIDE and MS-COCO.
Unless specified, F1@3 is taken to measure the performance.

\begin{table}[t]
    \centering
    \scalebox{0.93}{
        \small
        \begin{tabular}{p{62pt}<{\raggedright}|p{17pt}<{\centering}p{17pt}<{\centering}p{17pt}<{\centering}|p{17pt}<{\centering}p{17pt}<{\centering}p{17pt}<{\centering}}
        \multirow{2}{*}{\hspace{-5pt}Method} & \multicolumn{3}{c}{ZSL} & \multicolumn{3}{c}{GZSL}\\
        \cline{2-7}
        & \rule{0pt}{8pt}P@3 & R@3 & F1@3 & P@3 & R@3 & F1@3\\
        \shline
        \hspace{-5pt}\rule{0pt}{7pt}Deep0Tag~\cite{rahman2019deep} & 26.5 & 65.9 & 37.8 & 43.2 & 52.2 & 47.3 \\
        \hspace{-5pt}SDL$_{\text{M2}}$~\cite{ben2021semantic} & 26.3 & 65.3 & 37.5 & 59.0 & 60.8 & 59.9 \\
        \hspace{-5pt}CLIP~\cite{radford2021learning} & 30.5 & 76.0 & 43.5 & 31.7 & 37.1 & 34.2 \\
        \hspace{-5pt}DualCoOp~\cite{sun2022dualcoop} & 35.3 & 87.6 & 50.3 & 58.4 & 68.1 & 62.9 \\
        \hspace{-5pt}DualCoOp++~\cite{hu2023dualcoop++} & \underline{36.8} & \underline{91.4} & \underline{52.5} & \underline{59.4} & \underline{69.3} & \underline{64.0} \\
        \rowcolor{Gray}\hspace{-5pt}RAM (\textbf{ours}) & \textbf{38.2} & \textbf{95.0} & \textbf{54.5} & \textbf{62.5} & \textbf{72.9} & \textbf{67.3} \\
        \rowcolor{Gray}\hspace{-5pt}\textbf{$\Delta$ Prev. Best} & \textcolor{red}{\textbf{+1.4}} & \textcolor{red}{\textbf{+3.6}} & \textcolor{red}{\textbf{+2.0}} & \textcolor{red}{\textbf{+3.1}} & \textcolor{red}{\textbf{+3.6}} & \textcolor{red}{\textbf{+3.3}} \\
        \end{tabular}
    }
    \vspace{-0.25cm}
    \caption{\textbf{Results (\%)} on the open-vocabulary MS-COCO.}
    \label{tab:main_coco}
    \vspace{-0.3cm}
\end{table}

\begin{table}[t]
    \centering
    \scalebox{0.93}{
        \small
        \begin{tabular}{p{62pt}<{\raggedright}|p{17pt}<{\centering}p{17pt}<{\centering}p{17pt}<{\centering}|p{17pt}<{\centering}p{17pt}<{\centering}p{17pt}<{\centering}}
        \multirow{2}{*}{\hspace{-5pt}Method} & \multicolumn{3}{c}{RAP-OV} & \multicolumn{3}{c}{PA100K-OV}\\
        \cline{2-7}
        & \rule{0pt}{8pt}F1@3 & F1@5 & mAP & F1@3 & F1@5 & mAP\\
        \shline
        \hspace{-5pt}\rule{0pt}{7pt}BiAM~\cite{narayan2021discriminative} & 19.3 & 23.0 & 17.3 & 9.6 & 27.6 & 22.5 \\
        \hspace{-5pt}CLIP~\cite{radford2021learning} & 15.1 & 21.9 & 19.1 & 11.2 & 19.7 & 24.0 \\
        \hspace{-5pt}DualCoOp~\cite{sun2022dualcoop} & \underline{37.6} & \underline{56.0} & 36.3 & 33.5 & 50.4 & 39.1 \\
        \hspace{-5pt}ML-Decoder~\cite{ridnik2023ml} & 37.0 & 50.3 & \underline{39.8} & \underline{51.1} & \underline{65.4} & \underline{43.2} \\
        \hspace{-5pt}MKT~\cite{he2023open} & 19.0 & 27.6 & 24.1 & 18.7 & 20.9 & 27.0 \\
        \rowcolor{Gray}\hspace{-5pt}RAM (\textbf{ours}) & \textbf{48.9} & \textbf{66.1} & \textbf{53.5} & \textbf{66.4} & \textbf{81.5} & \textbf{55.0} \\
        \rowcolor{Gray}\hspace{-5pt}\textbf{$\Delta$ Prev. Best} & \textcolor{red}{\textbf{+11.3}} & \textcolor{red}{\textbf{+10.1}} & \textcolor{red}{\textbf{+13.7}} & \textcolor{red}{\textbf{+15.3}} & \textcolor{red}{\textbf{+16.1}} & \textcolor{red}{\textbf{+11.8}} \\
        \end{tabular}
    }
    \vspace{-0.25cm}
    \caption{\textbf{Results (\%)} on open-vocabulary Pedestrian Attribute Recognition. We take GZSL to construct OV benchmarks.}
    \label{tab:main_par}
    \vspace{-0.3cm}
\end{table}

\begin{table}[t]
    \centering
    \scalebox{0.93}{
        \small
        \begin{tabular}{p{62pt}<{\raggedright}|p{17pt}<{\centering}p{17pt}<{\centering}p{17pt}<{\centering}|p{17pt}<{\centering}p{17pt}<{\centering}p{17pt}<{\centering}}
        \multirow{2}{*}{\hspace{-5pt}Method} & \multicolumn{3}{c}{MultiScene-OV} & \multicolumn{3}{c}{MLRSNet-OV}\\
        \cline{2-7}
        & \rule{0pt}{8pt}F1@3 & F1@5 & mAP & F1@3 & F1@5 & mAP\\
        \shline
        \hspace{-5pt}\rule{0pt}{7pt}BiAM~\cite{narayan2021discriminative} & 14.4 & 18.4 & 13.7 & 13.5 & 13.3 & 12.6 \\
        \hspace{-5pt}CLIP~\cite{radford2021learning} & 30.9 & 33.6 & 42.9 & 19.7 & 21.7 & 41.4 \\
        \hspace{-5pt}RemoteCLIP~\cite{liu2024remoteclip} & 27.3 & 29.5 & 33.4 & 27.3 & 30.6 & \underline{44.7} \\
        \hspace{-5pt}DualCoOp~\cite{sun2022dualcoop} & 35.8 & 37.6 & 24.5 & 37.0 & \underline{42.5} & 28.5 \\
        \hspace{-5pt}ML-Decoder~\cite{ridnik2023ml} & \underline{37.2} & 37.6 & 31.9 & \underline{37.3} & 40.2 & 34.8 \\
        \hspace{-5pt}MKT~\cite{he2023open} & 37.1 & \underline{38.2} & \underline{43.6} & 26.4 & 28.8 & 42.5 \\
        \rowcolor{Gray}\hspace{-5pt}RAM (\textbf{ours}) & \textbf{45.4} & \textbf{47.4} & \textbf{51.2} & \textbf{46.9} & \textbf{54.4} & \textbf{51.0} \\
        \rowcolor{Gray}\hspace{-5pt}\textbf{$\Delta$ Prev. Best} & \textcolor{red}{\textbf{+8.2}} & \textcolor{red}{\textbf{+9.2}} & \textcolor{red}{\textbf{+7.6}} & \textcolor{red}{\textbf{+9.6}} & \textcolor{red}{\textbf{+11.9}} & \textcolor{red}{\textbf{+6.3}} \\
        \end{tabular}
    }
    \vspace{-0.25cm}
    \caption{\textbf{Results (\%)} on open-vocabulary Remote-Sensing Image Classification. We take GZSL to construct OV benchmarks.}
    \label{tab:main_rs}
    \vspace{-0.5cm}
\end{table}

\noindent \textbf{Ablation on the matching approaches.}
We first investigate the importance of matching.
For a fair comparison, we only replace the matching process in the network.
As shown in Table~\ref{tab:abl_match}, (a) directly applying OT~\cite{sinkhorn1967concerning} exhibits inferior performance on unseen classes with margins of 4.9\% and 2.3\% F1@3 on NUS-WIDE and COCO, respectively, which verifies the effectiveness of our TKT and LPD.
Compared to (b) bipartite matching~\cite{kuhn1955hungarian} where each label is associated with only one image region, both OT and our KCOT achieve better results.
This reason is that such one-to-one matching overlooks significant overlapping semantics for each label, resulting in biased alignments.
In contrast, KCOT \textit{emphasizes more meaningful image regions by performing one-to-many assignments}.
For (c) average pooling and (d) re-weighting that both neglect the matching process, 
our method achieves significant improvements, which verifies the significance of matching.

\begin{table}[t]
    \centering
    \scalebox{0.89}{
        \small
        \begin{tabular}{p{50pt}<{\raggedright}|p{39pt}<{\centering}p{39pt}<{\centering}|p{39pt}<{\centering}p{39pt}<{\centering}}
        \multirow{2}{*}{Matching} & \multicolumn{2}{c}{NUS} & \multicolumn{2}{c}{COCO}\\
        \cline{2-5}
        & \rule{0pt}{8pt}ZSL & GZSL & ZSL & GZSL \\
        \shline
        \hspace{-5pt}Bipartite~\cite{kuhn1955hungarian} & - & - & 51.9 \textcolor{mgreen}{(-2.6)} & 66.1 \textcolor{mgreen}{(-1.2)} \\
        \hspace{-5pt}OT~\cite{sinkhorn1967concerning} & 43.5 \textcolor{mgreen}{(-4.9)} & 22.3 \textcolor{mgreen}{(-1.2)} & 52.2 \textcolor{mgreen}{(-2.3)} & 66.8 \textcolor{mgreen}{(-0.5)} \\
        \rowcolor{Gray}\hspace{-5pt}KCOT & \multicolumn{1}{l}{\textbf{48.4}} & \multicolumn{1}{l|}{\textbf{23.5}} & \multicolumn{1}{l}{\textbf{54.5}} & \multicolumn{1}{l}{\textbf{67.3}} \\
        \hspace{-5pt}Average~\cite{he2023open} & 41.8 \textcolor{mgreen}{(-6.6)} & 22.0 \textcolor{mgreen}{(-1.5)} & 51.5 \textcolor{mgreen}{(-3.0)} & 65.5 \textcolor{mgreen}{(-1.8)} \\
        \hspace{-5pt}Re-weight~\cite{sun2022dualcoop} & 41.2 \textcolor{mgreen}{(-7.2)} & 18.2 \textcolor{mgreen}{(-5.3)} & 49.8 \textcolor{mgreen}{(-4.7)} & 63.6 \textcolor{mgreen}{(-3.7)} \\
        \end{tabular}
    }
    \vspace{-0.25cm}
    \caption{\textbf{Ablation (\%)} on the matching approaches. We only replace the matching part in the network.}
    \label{tab:abl_match}
    \vspace{-0.3cm}
\end{table}

\begin{table}[t]
    \centering
    \scalebox{0.88}{
        \small
        \begin{tabular}{p{19pt}<{\centering}p{19pt}<{\centering}|p{17pt}<{\centering}p{17pt}<{\centering}p{17pt}<{\centering}|p{17pt}<{\centering}p{18pt}<{\centering}|p{17pt}<{\centering}p{18pt}<{\centering}}
        \multicolumn{2}{c|}{LLA} & \multicolumn{3}{c|}{KCOT} & \multicolumn{2}{c}{NUS} & \multicolumn{2}{c}{COCO}\\
        \cline{6-9}
        \rule{0pt}{8pt}$\mathcal{F}_{SAA}$ & $\mathcal{F}_{TSS}$ & OT & LPD & TKT & ZSL & GZSL & ZSL & GZSL \\
        \shline
        & & & & & \rule{0pt}{7pt}40.6 & 19.4 & 50.6 & 60.8\\
        \checkmark & & & & & 41.0 & 20.1 & 51.2 & 62.3 \\
        & \checkmark & & & & 41.2 & 21.8 & 50.9 & 64.5 \\
        \checkmark & \checkmark & & & & 41.8 & 22.0 & 51.5 & 65.5 \\
        \checkmark & \checkmark & \checkmark & & & 43.5 & 22.3 & 52.2 & 66.8 \\
        \checkmark & \checkmark & \checkmark & \checkmark & & 47.1 & 22.9 & 53.9 & 67.0\\
        \checkmark & \checkmark & \checkmark & & \checkmark & 44.9 & 22.6 & 53.7 & 66.9\\
        \rowcolor{Gray}\checkmark & \checkmark & \checkmark & \checkmark & \checkmark & \textbf{48.4} & \textbf{23.5} & \textbf{54.5} & \textbf{67.3} \\
        \end{tabular}
    }
    \vspace{-0.25cm}
    \caption{\textbf{Ablation (\%)} on the proposed modules.}
    \label{tab:abl_module}
    \vspace{-0.3cm}
\end{table}

\begin{table}[t]
    \centering
    \scalebox{0.90}{
        \small
        \begin{tabular}{p{66pt}<{\raggedright}|p{17pt}<{\centering}p{18pt}<{\centering}p{20pt}<{\centering}|p{17pt}<{\centering}p{18pt}<{\centering}p{19pt}<{\centering}}
        \multirow{2}{*}{Matching} & \multicolumn{3}{c}{NUS} & \multicolumn{3}{c}{COCO}\\
        \cline{2-7}
        & \rule{0pt}{8pt}ZSL & GZSL & Mem. & ZSL & GZSL & Mem. \\
        \shline
        \hspace{-5pt}CoOp & 41.5 & 21.0 & 17.6G & 50.4 & 60.9 & \textcolor{mgreen}{\textbf{2.3G}} \\
        \hspace{-5pt}Ours + CoOp & 45.4 & 22.8 & 20.2G & 53.2 & 66.6 & 5.1G \\
        \rowcolor{Gray}\hspace{-5pt}\textbf{$\bm{\Delta}$} & \textcolor{red}{\textbf{+3.9}} & \textcolor{red}{\textbf{+1.8}} & - & \textcolor{red}{\textbf{+2.8}} & \textcolor{red}{\textbf{+5.7}} & - \\
        \cline{1-7}
        \hspace{-5pt}DualCoOp$^{*}$ & 43.2 & 21.7 & $>$24G & 51.3 & 62.2 & 3.0G \\
        \hspace{-5pt}Ours + DualCoOp$^{*}$ & 46.8 & \textbf{24.1} & $>$24G & 53.7 & \textbf{68.5} & 5.9G \\
        \rowcolor{Gray}\hspace{-5pt}\textbf{$\bm{\Delta}$} & \textcolor{red}{\textbf{+3.6}} & \textcolor{red}{\textbf{+2.4}} & - & \textcolor{red}{\textbf{+2.4}} & \textcolor{red}{\textbf{+6.3}} & -\\
        \cline{1-7}
        \hspace{-5pt}Ours & \textbf{48.4} & 23.5 & \textcolor{mgreen}{\textbf{8.5G}} & \textbf{54.5} & 67.3 & 4.4G \\
        \end{tabular}
    }
    \vspace{-0.25cm}
    \caption{\textbf{Ablation (\%)} on the label prompting. ``Mem.'' denotes the GPU memory usage reported with a batch size of 32. $*$ indicates the results based on our own reproduction.}
    \label{tab:abl_prompt}
    \vspace{-0.6cm}
\end{table}

\begin{figure*}[t]
    \centering
    \includegraphics[width=1.\linewidth]{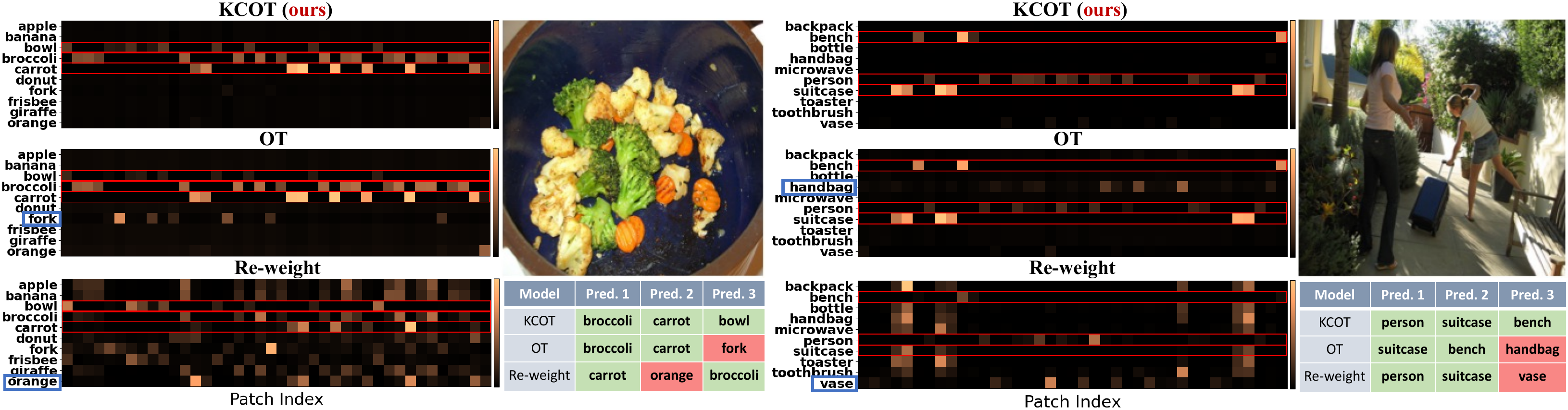}
    \vspace{-0.7cm}
     \caption{\textbf{Comparisons of different matching methods,} including independent re-weighting in~\cite{sun2022dualcoop}, OT~\cite{distances2013lightspeed} and our KCOT. The model is trained under different matching approaches. GT labels are marked in red boxes. Noisy matching leads to error-prone predictions, as shown in the table at the lower right corner. Correct and incorrect top-3 predictions are marked in green and red cells, respectively.}
	\label{fig:ot_glb}
    \vspace{-0.6cm}
\end{figure*}

\noindent \textbf{Effect of proposed modules.}
As shown in Figure~\ref{fig:intro_all} (c), both LLA and KCOT bring significant improvements.
Detailed ablations are shown in Table~\ref{tab:abl_module}.
(1) SAA enhances performance across multiple datasets in a \textit{parameter-free} manner, while TSS achieves significant improvements for GZSL task (+2.4\% and +3.7\% F1@3 on NUS-WIDE and COCO, respectively) due to its learning and selection ability. 
(2) The performance is remarkably improved when equipped with KCOT.
Notably, LPD achieves substantial improvements with margins of 3.6\% and 1.7\% F1@3 on NUS-WIDE and COCO.
This is due to that the unbalanced marginal $\bm{\tilde{u}}$ in LPD effectively reduces the impact of noise from irrelevant regions, suppressing \textit{meaningless} matching.

\noindent \textbf{Ablation on the label prompting.}
In Table~\ref{tab:abl_prompt}, our approach can serve as a versatile framework that can be integrated with various label prompting methods, \eg, CoOp~\cite{zhou2022learning} and DualCoOp~\cite{sun2022dualcoop}. The results show that our method significantly enhances their performance. Additionally, our method requires less GPU memory than both CoOp and DualCoOp, which suffer from huge memory consumption when dealing with large-scale categories like NUS-WIDE.

\begin{figure}[t]
    \centering
    \begin{minipage}[t]{0.23\textwidth}
    \centering
    \includegraphics[width=1\linewidth]{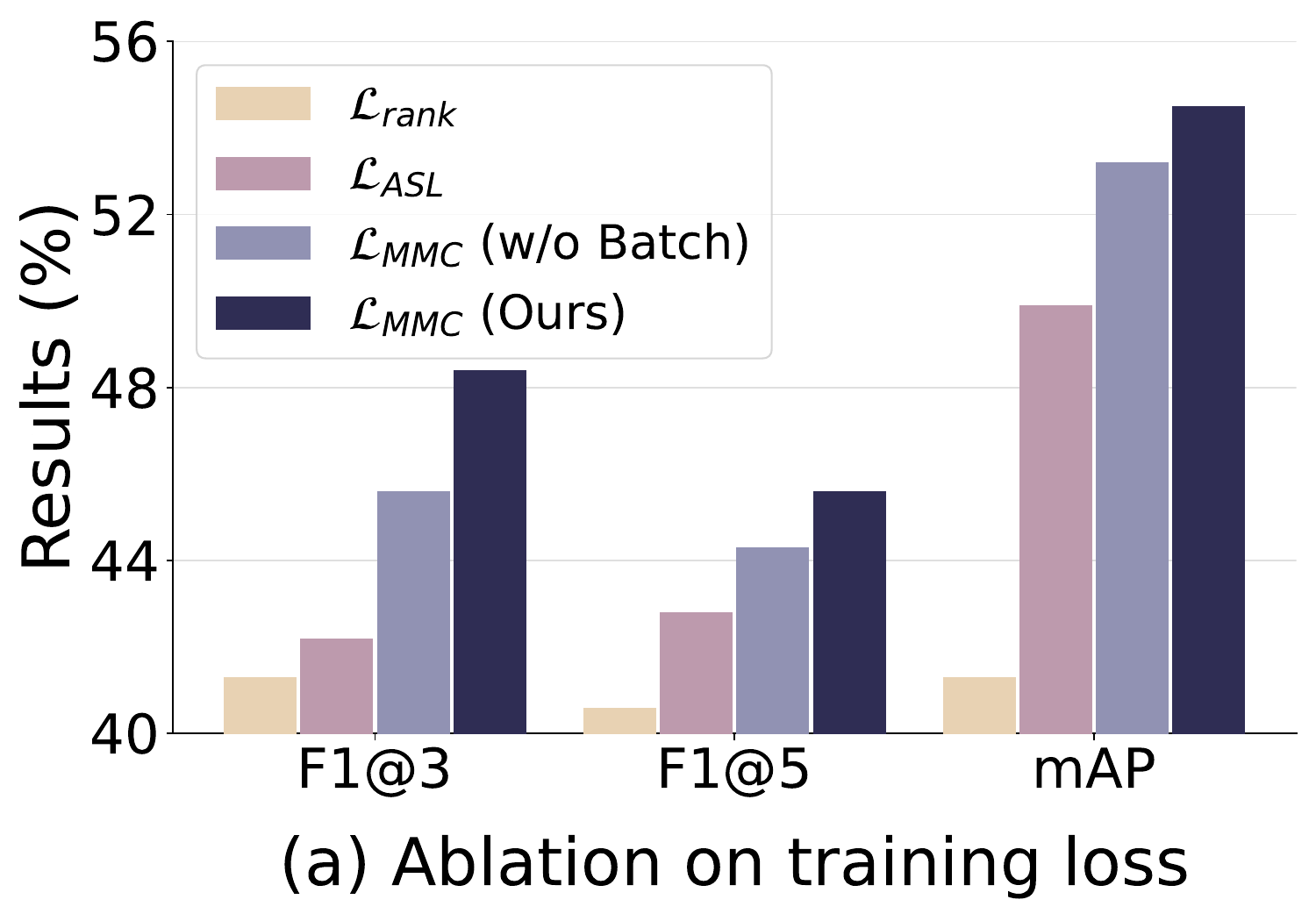}
    \end{minipage}
    \begin{minipage}[t]{0.23\textwidth}
    \centering
    \includegraphics[width=1\linewidth]{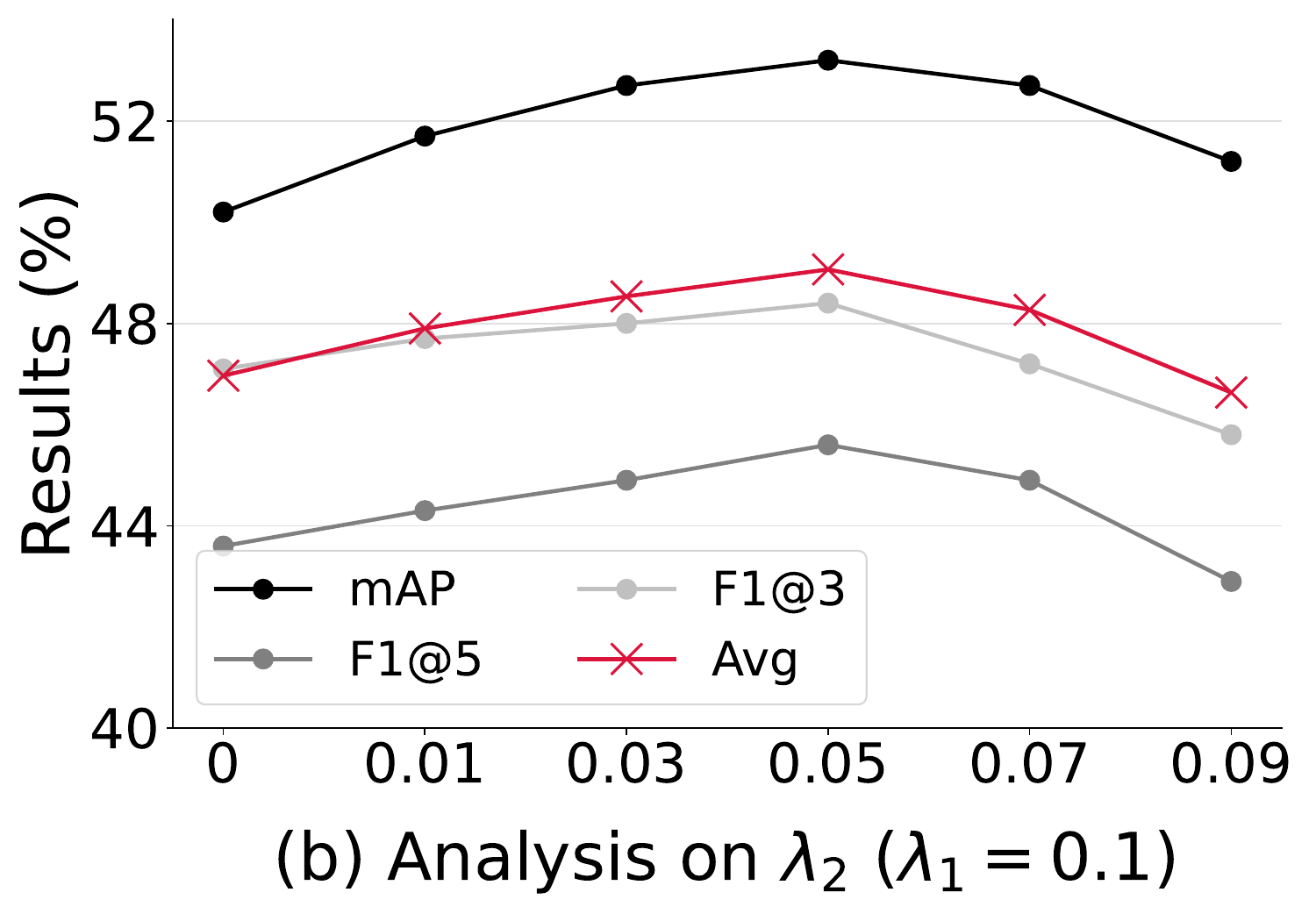}
    \end{minipage}
    \vspace{-0.3cm}
    \caption{(a) \textbf{Ablation (\%)} on training loss, including ranking loss~\cite{gong2013deep}, ASL~\cite{ridnik2021asymmetric} and our MMC loss. (b) \textbf{Analysis (\%)} of parameter $\lambda$.}
    \label{fig:ana}
    \vspace{-0.5cm}
\end{figure}

\noindent \textbf{Ablation on the training loss.}
As shown in Figure~\ref{fig:ana} (a), on NUS-ZSL, MMC loss exhibits notable advantages to ASL~\cite{ridnik2021asymmetric} and ranking loss~\cite{gong2013deep}.
The batch operation brings further improvements by increasing negative references in MMC.
In \textit{supplementary material}, we reveal that MMC loss is more effective for large-scale categories (\eg, NUS-WIDE), due to its contrastive paradigm.

\subsection{Further Analyses}
\begin{figure}[t]
    \centering
    \includegraphics[width=0.84\linewidth]{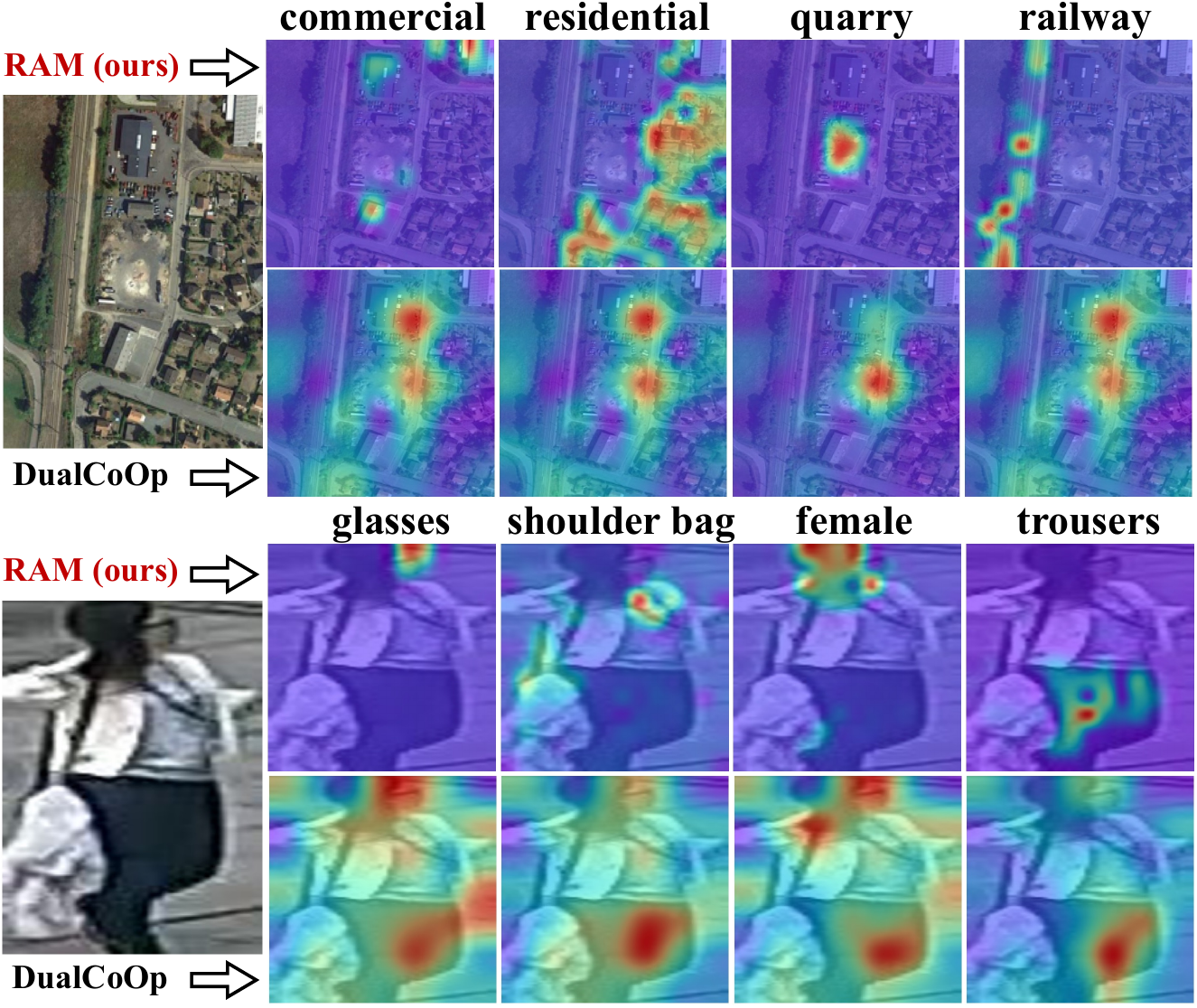}
    \vspace{-0.2cm}
     \caption{\textbf{Visualizations of matching results} with recovery (our RAM) and without recovery (DualCoOp) in specific domains.
     }
	\label{fig:ot_ind}
    \vspace{-0.5cm}
\end{figure}

\noindent \textbf{Qualitative analysis of recovery results.}
In Figure~\ref{fig:ot_ind}, our method effectively constructs fine-grained local features, while overlooking the loss of locality~\cite{sun2022dualcoop} leads to noisy regional features and illusory matching.

\noindent \textbf{Qualitative analysis of matching results.}
In Figure~\ref{fig:ot_glb}, re-weighting distributes high-response regions to most labels, leading to incorrect predictions.
OT alleviates such misaligned matching while still affected by some noisy regions.
By contrast, our KCOT focuses more precisely on the target classes, yielding accurate predictions.

\noindent \textbf{Sensitivity analysis of parameter $\lambda$.}
We fix $\lambda_1=0.1$ as a common choice~\cite{chen2022plot, zhu2024awt} and analyze $\lambda_2$ which controls the intensity of TKT.
As shown in Figure~\ref{fig:ana} (b), RAM gains improvements across a wide range of $\lambda_2$ (\ie, from 0 to 0.07).
Besides, large values of $\lambda_2$ should be avoided.

\section{Conclusion}
\label{sec:con}
This paper presents RAM, a simple yet effective framework that enhances OVMLR performance by addressing two significant limitations.
We reveal that regional semantics and accurate matching are essential to OVMLR, which are thus resolved by our proposed LLA and KCOT.
Extensive experiments on six datasets from three distinct domains demonstrate the state-of-the-art performance of RAM.

\noindent \textbf{Limitations and broader impacts.}
In this work, we only explored our method on OVMLR.
However, RAM can be extended to segmentations, which will be our future work.
Moreover, the implications on individual privacy should be carefully considered when applied to pedestrian domain.

\section*{Acknowledgments}
This work was supported by the Beijing Natural Science Foundation JQ23016, the Chinese National Natural Science Foundation Projects 62476273, the Science and Technology Development Fund of Macau Project 0123/2022/A3, 0044/2024/AGJ and InnoHK program.

\raggedbottom
{
    \small
    \bibliographystyle{ieeenat_fullname}
    \bibliography{main}
}

\clearpage
\appendix

\section{Additional Illustration on Method}

\subsection{Derivation of KCOT}
\label{app:deriv_kcot}
This section gives a detailed derivation of the KCOT problem in Sec.4.3.
The KCOT problem is formulated as:
\begin{equation}
\begin{aligned}
    &\mathop{\min}_{\bm{P}} \sum_{k=1}^{M} \sum_{i=1}^{N} \bm{P}_{ki} \bm{C}_{ki}  - \lambda_1 H(\bm{P}) + \lambda_2 D_{KL}(\bm{P}||\bm{\widetilde{P}}), \\
    &\ \text{s.t.}\ \ \bm{P} \bm{1}^N=\bm{\tilde{u}},\ \bm{P}^\mathsf{T} \bm{1}^M=\bm{v}.
\end{aligned}
\end{equation}
Based on the definition of KL Divergence and entropy, the minimization objective can be simplified as follows:
\begin{equation}
\begin{aligned}
    &\sum_{k=1}^{M} \sum_{i=1}^{N} \bm{P}_{ki} \bm{C}_{ki}  - \lambda_1 H(\bm{P}) + \lambda_2 D_{KL}(\bm{P}||\bm{\widetilde{P}}), \\
    =&\sum_{k=1}^{M} \sum_{i=1}^{N} \bm{P}_{ki} \bm{C}_{ki}  - \lambda_1 H(\bm{P}) + \lambda_2 \sum_{k=1}^{M} \sum_{i=1}^{N}(\bm{P}_{ki}\log \frac{\bm{P}_{ki}}{\bm{\widetilde{P}}_{ki}}), \\
    =&\sum_{k=1}^{M} \sum_{i=1}^{N} \bm{P}_{ki} \bm{C}_{ki}  - \lambda_1 H(\bm{P}) + \lambda_2 \sum_{k=1}^{M} \sum_{i=1}^{N}(\bm{P}_{ki}\log \bm{P}_{ki}) \\
    &-\lambda_2 \sum_{k=1}^{M} \sum_{i=1}^{N}(\bm{P}_{ki}\log \bm{\widetilde{P}}_{ki}), \\
    =&\sum_{k=1}^{M} \sum_{i=1}^{N} \bm{P}_{ki}(\bm{C}_{ki}-\lambda_2\log \bm{\widetilde{P}}_{ki}) - (\lambda_1+\lambda_2) H(\bm{P}).\\
\end{aligned}
\end{equation}
Here, the cost is transformed into $\bm{\widetilde{C}}_{ki} = \bm{C}_{ki} - \lambda_2\log\bm{\widetilde{P}}_{ki}$ and the coefficient of entropy $H(\bm{P})$ is re-weighted into $\tilde{\lambda}=\lambda_1+\lambda_2$.
As a result, the KCOT problem is simplified to the following form:
\begin{equation}
\begin{aligned}
    &\mathop{\min}_{\bm{P}} \sum_{k=1}^{M} \sum_{i=1}^{N} \bm{P}_{ki} \bm{\widetilde{C}}_{ki} -\tilde{\lambda} H(\bm{P}), \\
    &\ \text{s.t.}\ \ \bm{P} \bm{1}^N=\bm{\tilde{u}},\ \bm{P}^\mathsf{T} \bm{1}^M=\bm{v}.
\end{aligned}
\label{eq:kcot_final_supp}
\end{equation}
Eq.~\ref{eq:kcot_final_supp} is equivalent to an entropic OT problem, and can be solved by Sinkhorn algorithm~\cite{sinkhorn1967concerning} within several iterations.
Compared to OT, our KCOT enhances performance without increasing the problem complexity.
Moreover, the solution of KCOT does not involve any gradient propagation or updates on the model parameters, which is highly efficient and transferable.

\subsection{Pseudo Code}
The procedure of KCOT is summarized in Alg.~\ref{alg:kcot}.
In our studies, $t_{max}=100$ ensures convergence in most cases, which exhibits great efficiency.

\begin{algorithm}[t]
\caption{Knowledge-Constrained Optimal Transport}
\label{alg:kcot}
\begin{algorithmic}[1] 
\REQUIRE ~~\\
Visual set $\bm{X}^{\psi}$ and label set $\bm{T}$;\\
Teacher plan $\bm{\widetilde{P}}$ and parameters $\lambda_1, \lambda_2$, $t_{max}$;\\
\ENSURE ~~\\
Optimal transport plan $\bm{P}^{*}$;
\STATE Calculate the cost matrix $\bm{C}$ with $\{\bm{X}^{\psi}, \bm{T}\}$ in Eq. 7;
\STATE Set unbalanced marginal distribution $\bm{\tilde{u}}$ for visual set according to Eq. 8. Set balanced marginal distribution $\bm{v}=\bm{1}^{N}/N$ for label set;
\STATE Calculate the transformed cost matrix $\bm{\widetilde{C}}$ in Eq. 11, and calculate the parameter $\tilde{\lambda}=\lambda_1+\lambda_2$;
\STATE Initialize $\bm{b}^0=\bm{1}$, iteration $t=0$;
\WHILE {$t\leq t_{max}$ and not converge}
\STATE $\bm{a}^t=\bm{\tilde{u}}/((\text{exp}(-\bm{\widetilde{C}}/\tilde{\lambda})\bm{b}^{t-1})$;
\STATE $\bm{b}^t=\bm{v}/((\text{exp}(-\bm{\widetilde{C}}/\tilde{\lambda})^{\mathsf{T}}\bm{a}^{t})$;
\ENDWHILE 
\STATE Transport plan $\bm{P}^{*} = \text{diag}(\bm{a}^t)\text{exp}(-\bm{\widetilde{C}}/\tilde{\lambda})\text{diag}(\bm{b}^t)$;
\RETURN $\bm{P}^{*}$;
\end{algorithmic}
\end{algorithm}

\subsection{Discussion on SAA (How does SAA work?)}
\label{app:dis_saa}
\noindent 
Suppose the input sequences are projected into $\{\bm{Q}, \bm{K}, \bm{V}\}$.
The attention score $a_{ij}$ is determined by $a_{ij}=\bm{Q}_i\bm{K}_j^{\mathsf{T}}$.
Due to the image-level objective, the pre-trained attention always focuses on \textit{dominant patch} with large attention scores.
Our SAA replaces the calculation with $\tilde{a}_{ij}=\bm{V}_i\bm{V}_j^{\mathsf{T}}$.
Since the angle between $V_i$ and $V_i$ is 0, with similar magnitudes,
$\tilde{a}_{ii}$ is always greater than $\tilde{a}_{ij}(j\neq i)$, producing diagonal-style attention maps, which ensures better focus on itself.

\subsection{Discussion on the matching framework}
\begin{figure}[t]
    \centering
    \includegraphics[width=0.98\linewidth]{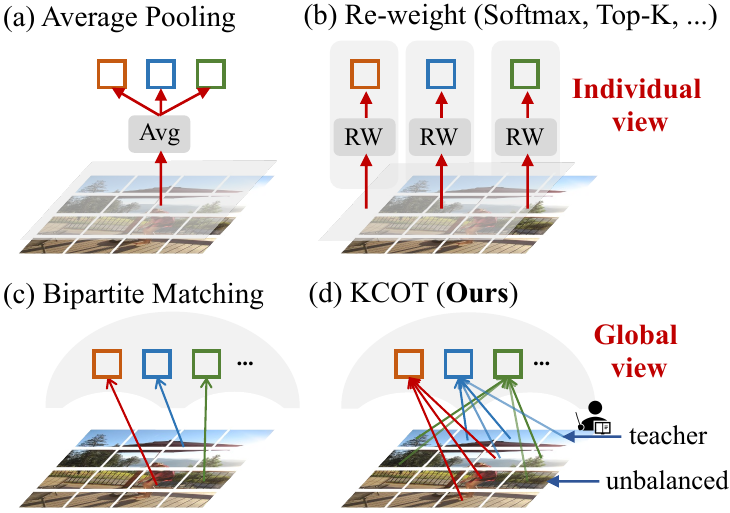}
     \caption{Matching comparisons. (a) Simple average pooling in~\cite{he2023open, zhu2024query} and (b) independent re-weighting in~\cite{sun2022dualcoop, hu2023dualcoop++, huynh2020shared,liu2021query2label,ma2024text} overlook the matching property, assigning features to each label individually.
     (c) Bipartite matching~\cite{kuhn1955hungarian} finds one-to-one matching.
     (d) Our KCOT performs one-to-many matching under the guidance of teacher plan and unbalanced marginal.
     KCOT finds matching from a global view, suppressing meaningless matching.
     }
	\label{fig:intro_framework}
\end{figure}

As shown in Figure~\ref{fig:intro_framework}, 
the region aggregations of existing methods can be summarized into two types:
(a) Average pooling which distributes equal weights to all regions, involving predictions from most irrelevant areas.
(b) Independent re-weighting, which can be categorized into two clusters:
(\romannumeral1) Softmax re-weighting in~\cite{sun2022dualcoop, hu2023dualcoop++}, which normalizes the image-text similarities for each label individually.
(\romannumeral2) Attention re-weighting, which is a core strategy in traditional ML-ZSL methods~\cite{huynh2020shared, narayan2021discriminative}.
Recently, some works~\cite{liu2021query2label,ma2024text} have reused the ideas, which perform cross-attention between label embeddings and regional features.
However, the ``matching'' claimed in~\cite{liu2021query2label, ma2024text} is actually an independent re-weighting strategy, since the attention scores are calculated for each label individually.
The similarities between regions and all labels are \textit{not} jointly compared, as a result, it always emphasizes particular regions for each label including those non-GTs,
resulting in noisy and error-prone region aggregations.

In this work, we reformulate the problem from the \textit{set matching} perspective, where \textbf{the matching weights between different elements of the two sets are jointly determined}.
As shown in Figure~\ref{fig:intro_framework} (c), 
a straightforward way is to find a one-to-one matching (\ie, bipartite matching) between image regions and labels, which can be resolved by Hungarian algorithm as in~\cite{carion2020end}.
However, the excessively sparse matching results in unsatisfactory performance.
In contrast, we introduce optimal transport theory and formulate the KCOT problem, which implicitly suppresses matching to irrelevant labels by jointly comparing similarities of all labels, largely enhancing open-vocabulary performance.

\subsection{Detailed Architecture of TSS}
The input features $\bm{X}_l \in \mathbb{R}^{M\times d}$ are first reshaped into feature maps $\bm{X}'_l \in \mathbb{R}^{H'\times W'\times d}$, where $H'\times W'=d$.
To aggregate local semantics from different receptive fields, the kernel sizes of two convolutions are set as $3\times 3$ and $1\times 1$, respectively.
To reduce the number of parameters and network complexity, the cross-attention between image and text features is parameter-free~\cite{guo2023calip}.
The outputs from two streams are concatenated and pooled.
Then a depth-wise convolution is applied and the sigmoid is performed to get the spatial masks.
Above all, TSS recovers the local semantics through the lens of convolutions and integrates text features to highlight salient image regions.

\begin{figure}[t]
    \centering
    \includegraphics[width=0.98\linewidth]{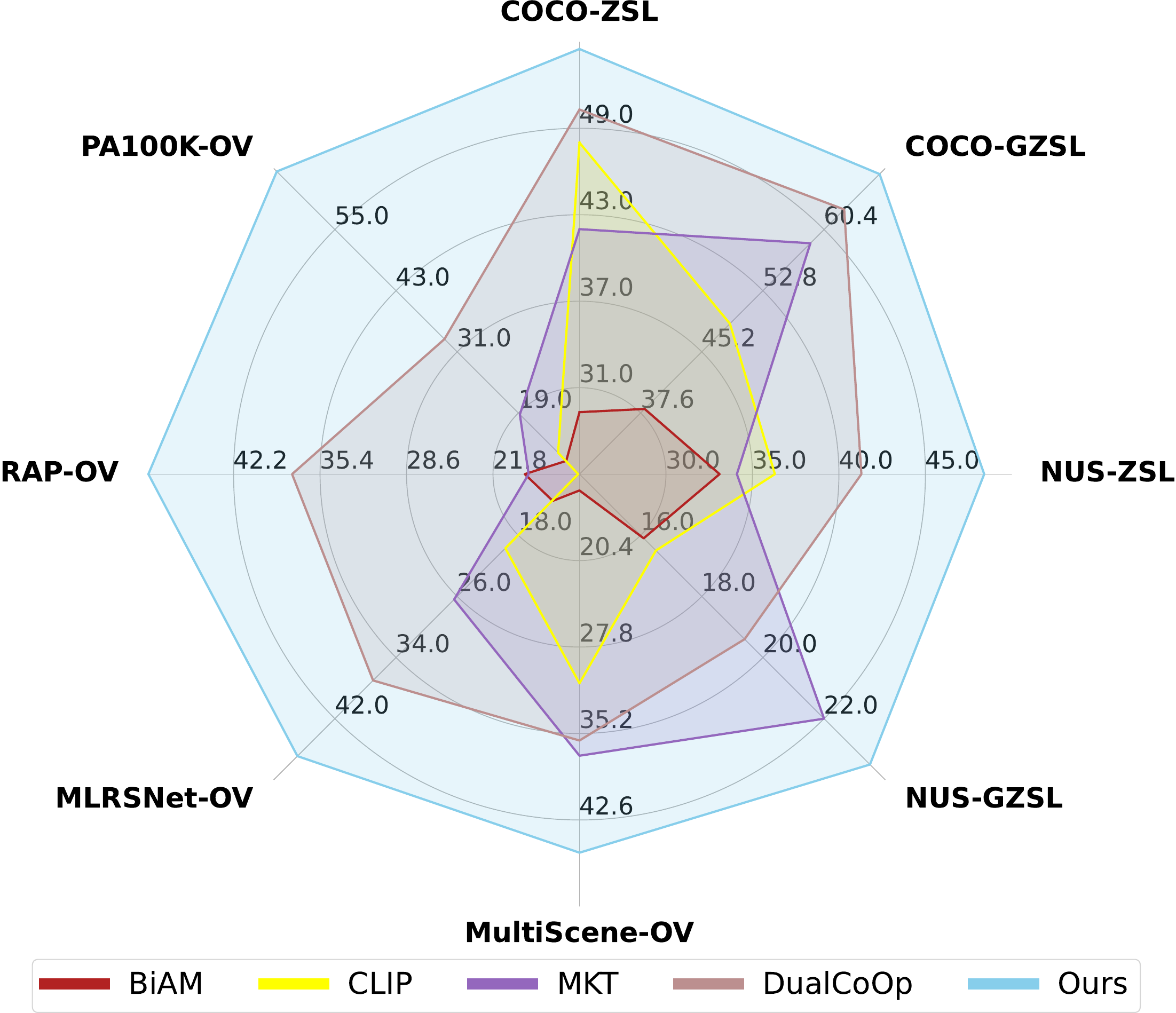}
     \caption{Performance comparisons to state-of-the-art methods on six datasets across different domains.
     }
	\label{fig:radar}
\end{figure}

\subsection{Discussion on the MMC loss}
We discuss the difference between our MMC loss and the commonly used ASL~\cite{ridnik2021asymmetric} and ranking loss~\cite{gong2013deep}.
For a mini-batch images $\{\bm{I}_b\}_{b=1}^{B}$, the predicted logit of the $i^th$ label for image $\bm{I}_b$ is denoted as $s(b,i)$.
Suppose $\mathcal{P}$ denotes the positive label set for a \textbf{single image}, and $\mathcal{P}_{B}$ is the positive set for the \textbf{mini-batch images}.
ASL is defined as:
\begin{small}
    \begin{equation}
        \left.\mathcal{L}_{ASL}\!=\!\frac{1}{B}\!\frac{1}{C}\!\sum_{b=1}^{B}\!\sum_{i=1}^{C}\!\left\{\begin{aligned}\!(&1\!-\!s(b,i))^{\gamma^+}\!\log (s(b,i)),\; i\!\in\!\mathcal{P},\\ 
        \!(&s(b,i))^{\gamma^-}\!\log(1\!-\!s(b,i)),\; i\!\notin\!\mathcal{P},
    \end{aligned}\right.\right.
    \label{eq:asl}
    \end{equation}
\end{small}
where $\gamma^+$ and $\gamma^-$ are two hyper-parameters to enable the asymmetric focus for positive and negative labels.
It is worth noting that ASL treats each label individually.
Although this can perform well in closed-set multi-label recognition, it is hard to generalize well to unseen classes.

Alternatively, contrastive learning is a promising paradigm to improve zero-shot generalization.
Ranking loss is a representative approach, which is defined as:
\begin{equation}
    \mathcal{L}_{rank}=\sum_{b=1}^{B} \sum_{\substack{i\in \mathcal{P} \\ j\notin \mathcal{P}}} \max (1+s(b,j)-s(b,i), 0).
\label{eq:rank}
\end{equation}
Different from ASL, ranking loss jointly compares all labels, learning the ordering relationships among these categories.
However, it still encounters two problems:
(1) the contrastive learning is restricted to a single sample, which hinders the generalization.
(2) the performance on large-scale categories is limited (as shown in Table~\ref{tab:add_abl_mmc}), possibly due to the unstable gradients caused by large values in loss.

In contrast, InfoNCE~\cite{oord2018representation} exhibits more stable optimization due to the normalized outputs.
Inspired by Supervised Contrastive Learning (SCL)~\cite{khosla2020supervised}, InfoNCE can be extended to multi-label scenarios by defining multiple positive pairs:
\begin{equation}
    \mathcal{L}\!=\!-\frac{1}{|\mathcal{P}|}\!\sum_{b=1}^{B}\log \frac{\text{exp}(s(b,i)/\tau')}{\sum_{j=1}^{N}\!\text{exp}(s(b,j)/\tau')}.
    \label{eq:scl}
\end{equation}
In this work, we further extend the negative references into the mini-batch, yielding MMC loss:
\begin{small}
\begin{equation}
    \mathcal{L}_{\text{MMC}}\!=\!-\frac{1}{|\mathcal{P}_B|}\!\sum_{(b,i)\in \mathcal{P}_B}\!\!\!\!\log \frac{\text{exp}(s(b,i)/\tau')}{\sum_{b^{'}\!=1}^{B}\!\sum_{j=1}^{N}\!\text{exp}(s(b^{'}\!,j)/\tau')},
    \label{eq:mmc_supp}
\end{equation}
\end{small}
Compared to ASL, MMC loss inherits the advantages of contrastive learning, enabling discriminative representations.
Compared to ranking loss and InfoNCE, MMC loss allows more diverse entries to be negative references, facilitating generalizable alignments.

\section{Datasets and Implementation Details}
\label{app:details}
\noindent \textbf{Construction of OV benchmarks.}
For NUS-WIDE and MS-COCO, we follow previous works to sample the seen and unseen classes, which results in 925/81 seen and unseen classes for NUS-WIDE, 48/17 seen and unseen classes for MS-COCO.

\noindent For PAR benchmarks (\ie, RAPv1 and PA100K), we sample over 30\% classes as unseen based on the frequency.
To ensure the quantity of training samples, we select the most frequent classes as seen labels, which results in 35/16 seen and unseen classes for RAPv1, 16/10 seen and unseen classes for PA100K.

\noindent For RS benchmarks (\ie, MultiScene and MLRSNet), we randomly sample over 30\% classes as unseen, which results in 20/16 seen and unseen classes for MultiScene, 40/20 seen and unseen classes for MLRSNet.

\noindent \textbf{Dataset statistics.}
As summarized in Table~\ref{tab:stat}, the six datasets contain challenges from distinct domains.
For RAPv1, PA100K and MLRSNet, we take the official test set for evaluation.
For MultiScene dataset, we take the official MultiScene-Clean subset for evaluation which contains 7K manually-annotated images.

\begin{table}[t]
    \centering
    \scalebox{0.94}{
        \small
        \begin{tabular}{p{55pt}<{\raggedright}p{27pt}<{\centering}|p{30pt}<{\centering}p{30pt}<{\centering}p{47pt}<{\centering}}
        \multirow{2}{*}{\hspace{-5pt}Datasets} & \multirow{2}{*}{Domain} & \multirow{2}{*}{Training} & GZSL & Labels \\
        &&& Testing & (seen/unseen) \\
        \shline
        \hspace{-5pt}\rule{0pt}{7pt}NUS-WIDE & Natural & 143K & 59K & 925 / 81 \\
        \hspace{-5pt}MS-COCO & Natural & 44K & 6K & 48 / 17 \\
        \hspace{-5pt}RAP-OV & PAR & 25K & 8K & 35 / 16 \\
        \hspace{-5pt}PA100K-OV & PAR & 59K & 10K & 16 / 10 \\
        \hspace{-5pt}MultiScene-OV & RS & 24K & 7K & 20 / 16 \\
        \hspace{-5pt}MLRSNet-OV & RS & 11K & 21K & 40 / 20 \\
        \end{tabular}
    }
    \caption{Dataset statistics of the open-vocabulary benchmarks. ``PAR'' denotes Pedestrian Attribute Recognition. ``RS'' denotes Remote Sensing image classification.}
    \label{tab:stat}
\end{table}

\begin{table}[t]
    \centering
    \scalebox{0.94}{
        \small
        \begin{tabular}{p{50pt}<{\centering}p{35pt}<{\centering}|p{20pt}<{\centering}p{20pt}<{\centering}|p{20pt}<{\centering}p{20pt}<{\centering}}
        \multirow{2}{*}{Method} & \multirow{2}{*}{\# Prompts} & \multicolumn{2}{c}{NUS} & \multicolumn{2}{c}{COCO}\\
        \cline{3-6}
        && \rule{0pt}{8pt}ZSL & GZSL & ZSL & GZSL \\
        \shline
        w/o prompt & - & 41.2 & 20.3 & 52.7 & 66.7 \\
        Deep prompt & 2 & 46.4 & 22.9 & \textbf{54.5} & 67.1 \\
        \rowcolor{Gray}Deep prompt & 4 & \textbf{48.4} & \textbf{23.5} & \textbf{54.5} & 67.3 \\
        Deep prompt & 8 & 47.9 & 23.4 & 54.1 & \textbf{67.7} \\
        Deep prompt & 16 & 47.9 & 23.3 & 53.6 & 67.5 \\
        Deep prompt & 32 & 45.0 & 22.7 & 53.1 & 67.5 \\
        \end{tabular}
    }
    \caption{Analysis on the deep label prompting.}
    \label{tab:abl_dlp}
\end{table}

\noindent \textbf{Implementation details.}
We apply several augmentation strategies to training images, including random crop, random flip, gaussian blur and random erasing~\cite{zhong2020random}.
During testing, we only perform resize operation.
Learning rate is set as 5e-6 for natural images and 5e-5 for other domains using AdamW~\cite{loshchilov2017decoupled} optimizer, and decays with cosine policy.
SGD optimizer with learning rate of 1e-3 is set for all learnable prompts following a common practice~\cite{zhou2022learning, sun2022dualcoop}.
LLA is applied in the last few layers and visual prompts are integrated to modulate global feature.
On RS and PAR, $\lambda_2$ is set as 0.01 to neutralize the effect of frozen knowledge.

\begin{table}[t]
    \centering
    \scalebox{0.94}{
        \small
        \begin{tabular}{p{65pt}<{\raggedright}|p{20pt}<{\centering}p{20pt}<{\centering}|p{20pt}<{\centering}p{20pt}<{\centering}}
        \multirow{2}{*}{\hspace{-5pt}Loss Type} & \multicolumn{2}{c}{NUS} & \multicolumn{2}{c}{COCO}\\
        \cline{2-5}
        & \rule{0pt}{8pt}ZSL & GZSL & ZSL & GZSL \\
        \shline
        \hspace{-5pt}ASL & 43.2 & 22.4 & 53.3 & 66.9 \\
        \hspace{-5pt}Ranking & 41.3 & 20.8 & 53.5 & 66.4 \\
        \hspace{-5pt}MMC (w/o batch) & 45.6 & \textbf{24.2} & 53.8 & \textbf{67.7} \\
        \hspace{-5pt}MMC & \textbf{48.4} & 23.5 & \textbf{54.5} & 67.3 \\
        \end{tabular}
    }
    \caption{Additional ablations on MMC loss.}
    \label{tab:add_abl_mmc}
\end{table}

\begin{table}[t]
    \centering
    \scalebox{0.94}{
        \small
        \begin{tabular}{p{65pt}<{\raggedright}|p{20pt}<{\centering}p{20pt}<{\centering}|p{20pt}<{\centering}p{20pt}<{\centering}}
        \multirow{2}{*}{\hspace{-5pt}Method} & \multicolumn{2}{c}{NUS} & \multicolumn{2}{c}{COCO}\\
        \cline{2-5}
        & \rule{0pt}{8pt}ZSL & GZSL & ZSL & GZSL \\
        \shline
        \hspace{-5pt}RAM (w/o SAA) & 45.2 & 22.5 & 52.9 & 66.8 \\
        \hspace{-5pt}RAM (w/o TSS) & 47.6 & 23.2 & 53.6 & 67.0 \\
        \rowcolor{Gray}\hspace{-5pt}RAM & \textbf{48.4} & \textbf{23.5} & \textbf{54.5} & \textbf{67.3} \\
        \end{tabular}
    }
    \caption{Additional ablations on LLA.}
    \label{tab:add_abl_lla}
\end{table}

\begin{figure*}[t]
    \centering
    \begin{minipage}[t]{0.245\textwidth}
    \centering
    \includegraphics[width=1\linewidth]{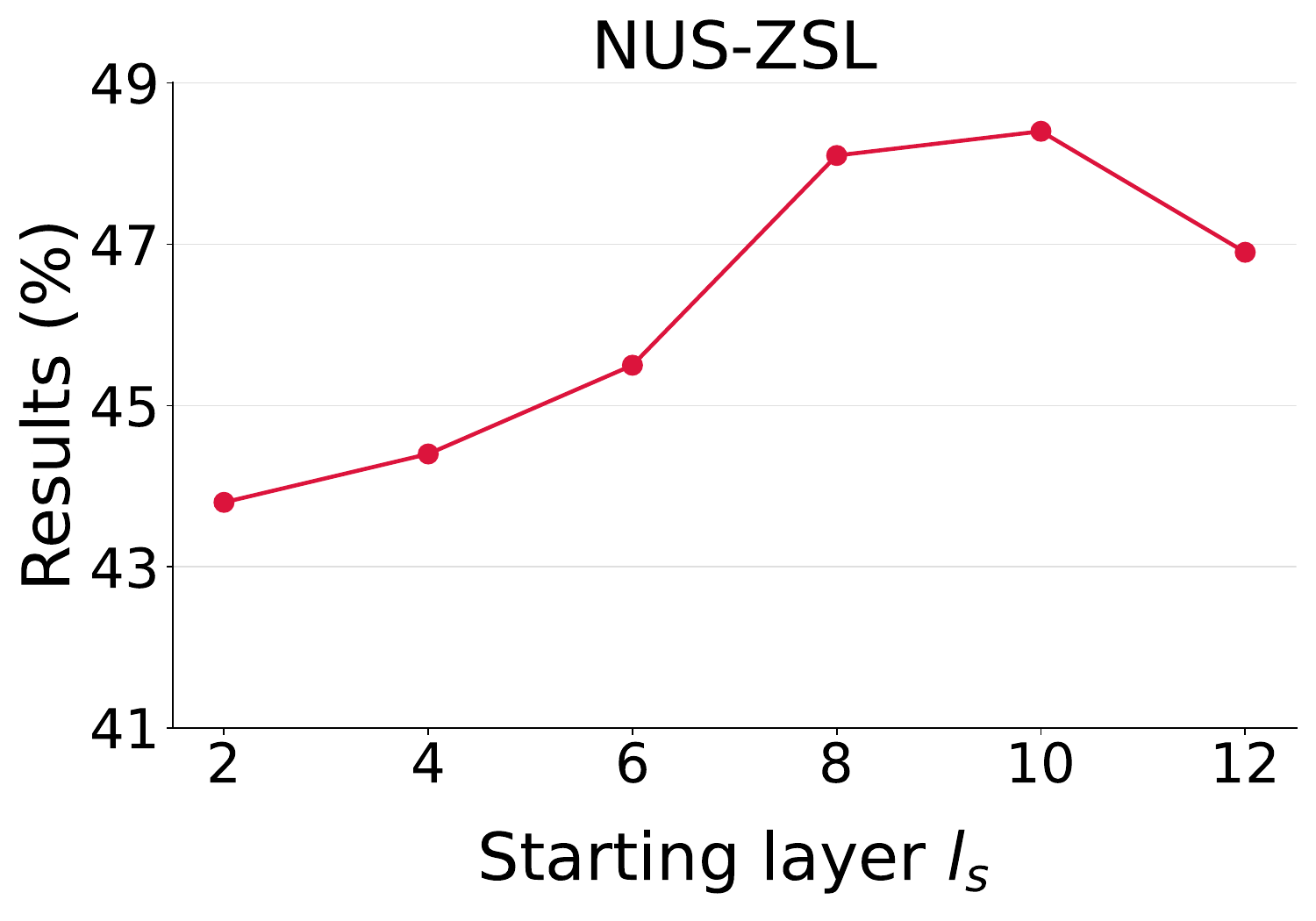}
    \end{minipage}
    \begin{minipage}[t]{0.245\textwidth}
    \centering
    \includegraphics[width=1\linewidth]{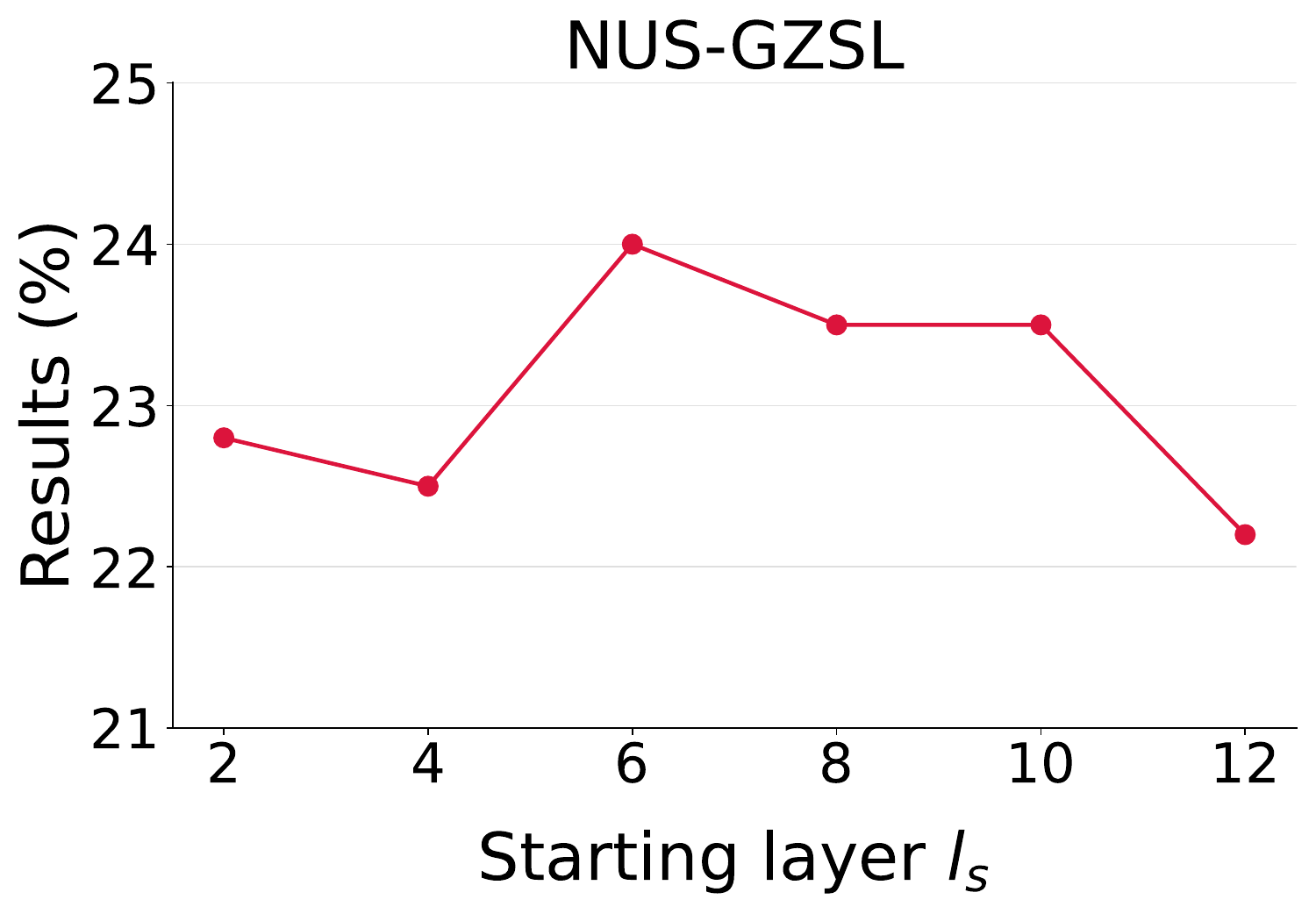}
    \end{minipage}
    \begin{minipage}[t]{0.245\textwidth}
    \centering
    \includegraphics[width=1\linewidth]{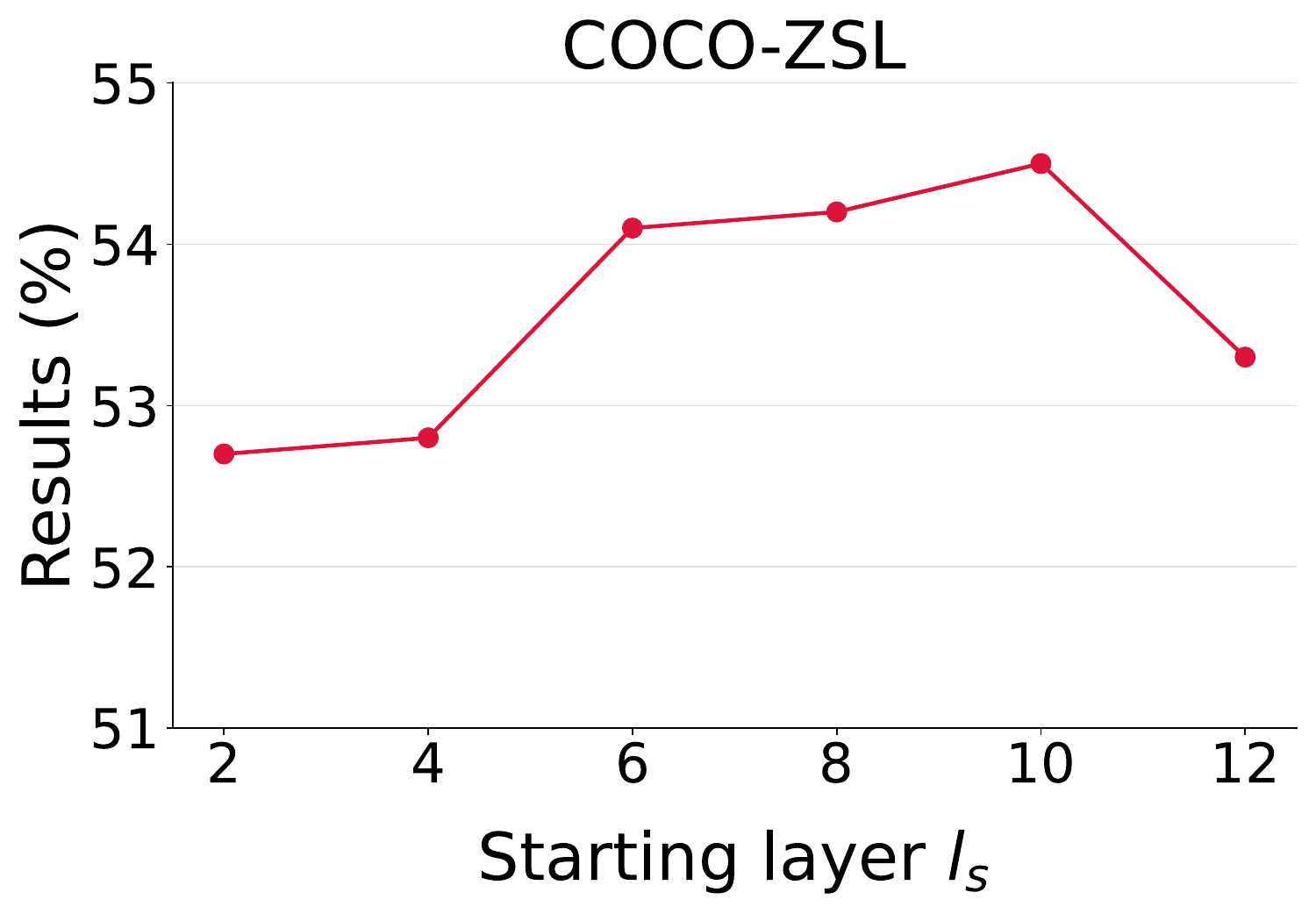}
    \end{minipage}
    \begin{minipage}[t]{0.245\textwidth}
    \centering
    \includegraphics[width=1\linewidth]{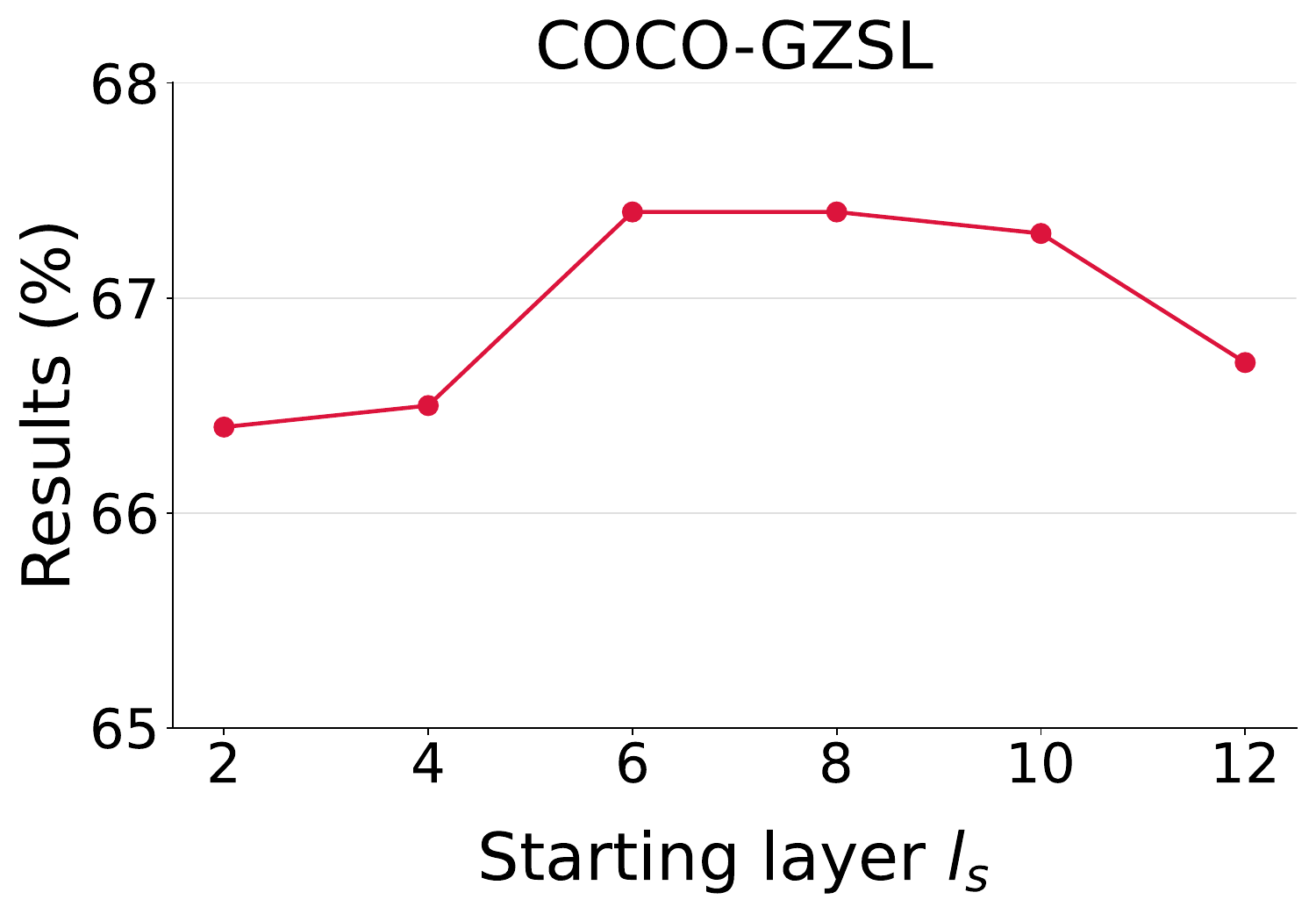}
    \end{minipage}
    \caption{Analysis on the starting layer $l_s$. F1@3 is reported on NUS-WIDE and MS-COCO. $l_s=10$ is a better trade-off.}
    \label{fig:start_layer}
\end{figure*}

\begin{table*}[t]
\vspace{-0.2cm}
    \centering
    \scalebox{0.94}{
        \small
        \begin{tabular}{p{90pt}<{\raggedright}|p{45pt}<{\centering}p{45pt}<{\centering}p{48pt}<{\centering}p{52pt}<{\centering}p{30pt}<{\centering}p{30pt}<{\centering}}
        \hspace{-5pt}Method & Train Speed (samples/s) & Test Speed (samples/s) & Train Mem. (Byte) & Test Mem. (Byte) & ZSL & GZSL \\
        \shline
        \hspace{-5pt}Baseline & 143.5 & 101.9 & 5.9G & 3.5G & 40.6 & 19.4 \\
        \hspace{-5pt}Baseline + LLA & 103.2 & 95.7 & 7.2G & 3.8G & 41.8 & 22.0 \\
        \rowcolor{Gray}\hspace{-5pt}\textbf{$\bm{\Delta}$ (by LLA)} & \textcolor{mgreen}{\textbf{$\downarrow$40.3}} & \textcolor{mgreen}{\textbf{$\downarrow$6.2 (6.1\%)}} & \textcolor{mgreen}{\textbf{$\uparrow$1.3G}} & \textcolor{mgreen}{\textbf{$\uparrow$0.3G (8.6\%)}} & \textcolor{red}{\textbf{$\uparrow$1.2}} & \textcolor{red}{\textbf{$\uparrow$2.6}} \\
        \cline{1-7}
        \rule{0pt}{8pt}\hspace{-5pt}Baseline + LLA + KCOT & 90.3 & 94.5 & 8.5G & 3.8G & 48.4 & 23.5 \\
        \rowcolor{Gray}\hspace{-5pt}\textbf{$\bm{\Delta}$ (by KCOT)} & \textcolor{mgreen}{\textbf{$\downarrow$12.9}} & \textcolor{mgreen}{\textbf{$\downarrow$1.2 (1.3\%)}} & \textcolor{mgreen}{\textbf{$\uparrow$1.3G}} & \textcolor{mgreen}{\textbf{$\uparrow$0.0G (0.0\%)}} & \textcolor{red}{\textbf{$\uparrow$6.6}} & \textcolor{red}{\textbf{$\uparrow$1.5}} \\
        \end{tabular}
    }
    \caption{Efficiency analysis on the proposed LLA and KCOT. ``Mem.'' is short for GPU memory usage. All statistics are obtained with a batch size of 32 on NUS-WIDE.}
    \label{tab:efficiency}
\end{table*}

\section{Additional Results}

\subsection{Analysis on the starting layer $l_s$}
As shown in Figure~\ref{fig:start_layer}, integrating LLA and textual prompts in early stages (\eg, $l_s<6$) leads to degraded performance, possibly due to overfitting.
$l_s=6$ achieves the best results on GZSL task while $l_s=10$ exhibits better performance on ZSL.
Overall, $l_s=10$ is a better trade-off and introduces fewer parameters and computational overhead.

\subsection{Analysis on the deep label prompting $Q_l$}
As shown in Table~\ref{tab:abl_dlp}, 
introducing textual prompts greatly improves performance.
A limited number of prompts (\eg, 2, 4 and 8) achieves great performance on both ZSL and GZSL.
Increasing the number of prompts (\eg, 16 and 32) does not bring further improvements.

\subsection{Additional ablations on the LLA} 
As shown in Table~\ref{tab:add_abl_lla}, we remove only SAA and TSS, respectively.
Notably, both SAA and TSS bring notable improvements and SAA is more important (\eg, with margins of 3.2\% and 1.6\% on NUS-ZSL and COCO-ZSL, respectively).
Note that SAA is parameter-free, which can be seamlessly integrated into more diverse tasks as an effective and efficient recovery.

\subsection{Additional ablations on the MMC loss}
In Table~\ref{tab:add_abl_mmc}, we provide additional results on both ZSL and GZSL tasks.
Notably, the proposed MMC loss achieves impressive improvements on NUS-WIDE, while the improvements on MS-COCO are less significant.
The reason is that:
with a larger number of categories, visually or semantically similar categories are more likely to overlap in the embedding space.
Compared to ASL, contrastive learning is more effective in handling fine-grained distinctions by explicitly pushing different labels apart in the embedding space.
Besides, removing batch operation in MMC loss yields better GZSL performance but significantly inferior ZSL performance.
We suggest tailoring the choice to specific task, as batch operations are easy to implement.

\subsection{Analysis on efficiency}
As presented in Table~\ref{tab:efficiency}, LLA brings certain computational overhead during training while remains lightweight during inference.
KCOT delivers remarkable performance gains with almost no extra overhead during inference (\ie, only 1.3\% speed drop and nearly 0.0\% memory increase), which verifies the efficiency of our proposed LLA and KCOT.

\subsection{Additional visualizations}
\begin{figure*}[t]
    \vspace{-0.3cm}
    \centering
    \includegraphics[width=0.9\linewidth]{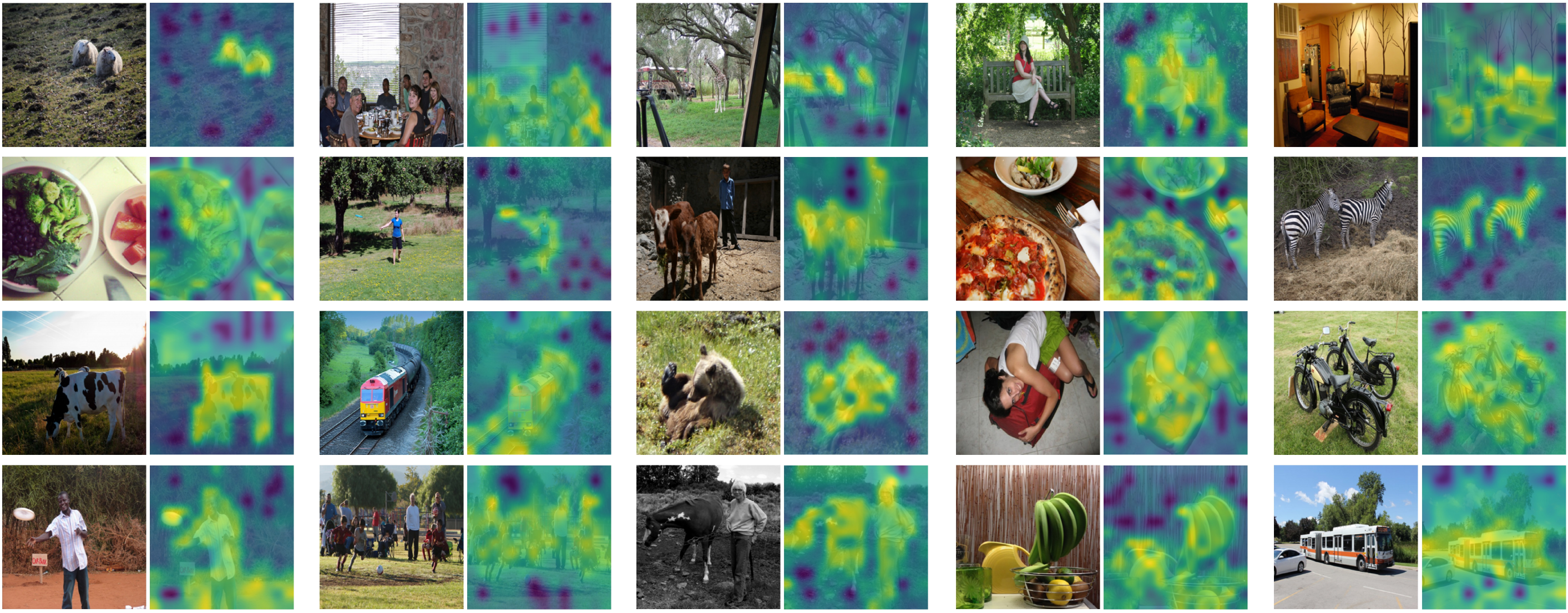}
    \vspace{-0.3cm}
     \caption{Visualizations of unbalanced marginal $\bm{\tilde{u}}$. Brighter color means higher weight. Best viewed in color.}
	\label{fig:supp_marginal}
    \vspace{-0.3cm}
\end{figure*}

\noindent \textbf{Unbalanced marginal $\bm{\tilde{u}}$.}
In Figure~\ref{fig:supp_marginal}, 
we visualize vector $\bm{\tilde{u}}\in \mathbb{R}^{M}$ in LPD by reshaping it into $\mathbb{R}^{H'\times W'}$ and resizing to the image size.
Note that this does not directly represent the weights of regions, but instead serves as a marginal constraint to the matching.
LPD successfully emphasizes foreground areas.
Notably, it avoids fully \textit{binarizing} the target regions, allowing most regions to remain partially highlighted, which ensures generalization to unseen classes.

\begin{figure*}[t]
    \centering
    \includegraphics[width=0.9\linewidth]{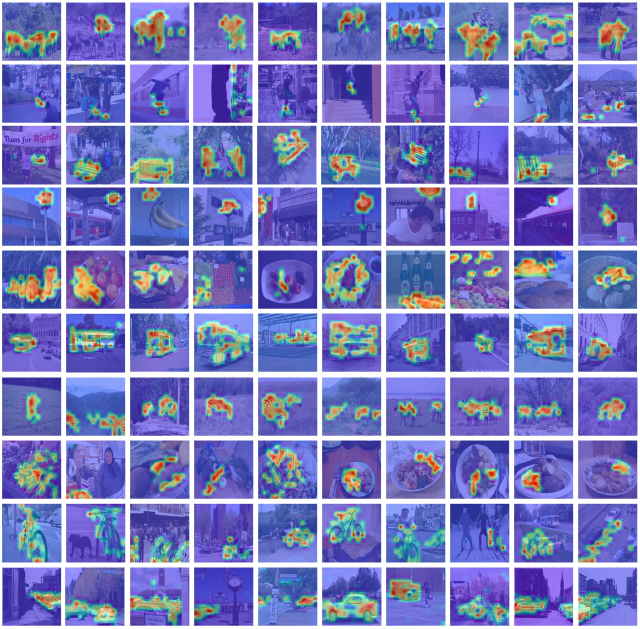}
    \vspace{-0.3cm}
     \caption{Visualizations of matching results (\textbf{individual view}). Each row is the matching weight for one label. From top to bottom are ``elephant'', ``skateboard'', ``bench'', ``clock'', ``banana'', ``bus'', ``zebra'', ``carrot'', ``bicycle'' and ``car'', respectively.}
	\label{fig:supp_ot_ind}
\end{figure*}

\noindent \textbf{Visualizations of matching results (individual view).}
As shown in Figure~\ref{fig:supp_ot_ind},
we visualize the matching results $\bm{P}^*\in \mathbb{R}^{M\times C}$ w.r.t. each label (\ie, $\mathbb{R}^{M}$, individual view).
We first reshape it into $\mathbb{R}^{H'\times W'}$ and resize to the image size.
Our method generates precise region-to-label matching under different circumstances, \eg, matching \textit{clock} in both indoor and outdoor, matching \textit{banana} of whole and sliced ones.
Notably, for a given category, our approach is capable of matching all corresponding objects in the image. 
For instance, it can find all \textit{elephants} when there are multiple in the scene, which is similar for \textit{banana}, \textit{car} and \textit{bicycle}.

\begin{figure*}[t]
    \vspace{-0.3cm}
    \centering
    \includegraphics[width=0.9\linewidth]{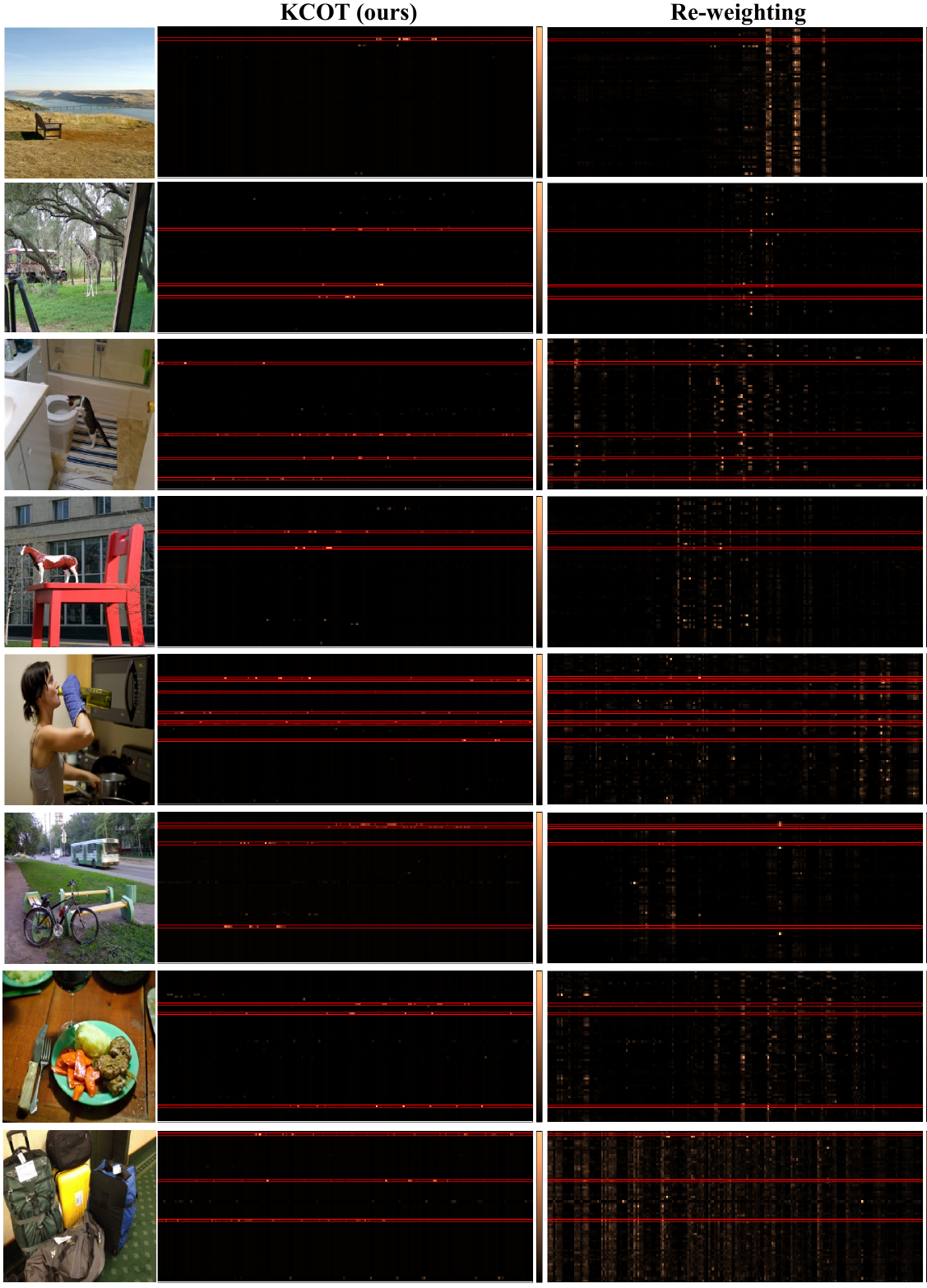}
    \vspace{-0.3cm}
     \caption{Visualizations of matching results (\textbf{global view}). The x-axis denotes the regions, and the y-axis corresponds to candidate labels. Ground-truth labels are marked in red boxes. Brighter color means higher matching weight.}
	\label{fig:supp_ot_glb}
\end{figure*}

\noindent \textbf{Visualizations of matching results (global view).}
In Figure~\ref{fig:supp_ot_glb},
we visualize the matching results $\bm{P}^*\in \mathbb{R}^{M\times C}$ w.r.t. all labels (\ie, global view).
We compare the proposed KCOT and independent re-weighting, which is widely used in previous works.
Re-weighting distributes high-response regions to most labels, which we refer as ``extraneous high-response'' phenomenon.
The reason is that it computes region weights for each label \textit{separately},
inevitably producing some regions with higher weights for every label.
In contrast, our KCOT successfully focus on the matching to target labels, facilitating precise and robust predictions.

\subsection{Performance comparison}
As shown in Figure~\ref{fig:radar}, our method surpasses state-of-the-art methods on six diverse datasets.
Notably, previous methods achieve limited performance in specialized domains such as PAR (\ie, PA100K-OV and RAP-OV) and RS (\ie, MultiScene-OV and MLRSNet-OV), while our method exhibits robust performance across different domains.

\end{document}

%% file: preamble.tex
\usepackage[dvipsnames]{xcolor}


%% file: arxiv.bbl
\begin{thebibliography}{65}
\providecommand{\natexlab}[1]{#1}
\providecommand{\url}[1]{\texttt{#1}}
\expandafter\ifx\csname urlstyle\endcsname\relax
  \providecommand{\doi}[1]{doi: #1}\else
  \providecommand{\doi}{doi: \begingroup \urlstyle{rm}\Url}\fi

\bibitem[Alvarez-Melis and Fusi(2020)]{alvarez2020geometric}
David Alvarez-Melis and Nicolo Fusi.
\newblock Geometric dataset distances via optimal transport.
\newblock \emph{Advances in Neural Information Processing Systems}, 33:\penalty0 21428--21439, 2020.

\bibitem[Bansal et~al.(2018)Bansal, Sikka, Sharma, Chellappa, and Divakaran]{bansal2018zero}
Ankan Bansal, Karan Sikka, Gaurav Sharma, Rama Chellappa, and Ajay Divakaran.
\newblock Zero-shot object detection.
\newblock In \emph{Proceedings of the European conference on computer vision (ECCV)}, pages 384--400, 2018.

\bibitem[Ben-Cohen et~al.(2021)Ben-Cohen, Zamir, Ben-Baruch, Friedman, and Zelnik-Manor]{ben2021semantic}
Avi Ben-Cohen, Nadav Zamir, Emanuel Ben-Baruch, Itamar Friedman, and Lihi Zelnik-Manor.
\newblock Semantic diversity learning for zero-shot multi-label classification.
\newblock In \emph{Proceedings of the IEEE/CVF International Conference on Computer Vision}, pages 640--650, 2021.

\bibitem[Carion et~al.(2020)Carion, Massa, Synnaeve, Usunier, Kirillov, and Zagoruyko]{carion2020end}
Nicolas Carion, Francisco Massa, Gabriel Synnaeve, Nicolas Usunier, Alexander Kirillov, and Sergey Zagoruyko.
\newblock End-to-end object detection with transformers.
\newblock In \emph{European conference on computer vision}, pages 213--229. Springer, 2020.

\bibitem[Chen et~al.(2022)Chen, Yao, Song, Li, Rao, and Zhang]{chen2022plot}
Guangyi Chen, Weiran Yao, Xiangchen Song, Xinyue Li, Yongming Rao, and Kun Zhang.
\newblock Plot: Prompt learning with optimal transport for vision-language models.
\newblock \emph{arXiv preprint arXiv:2210.01253}, 2022.

\bibitem[Chua et~al.(2009)Chua, Tang, Hong, Li, Luo, and Zheng]{chua2009nus}
Tat-Seng Chua, Jinhui Tang, Richang Hong, Haojie Li, Zhiping Luo, and Yantao Zheng.
\newblock Nus-wide: a real-world web image database from national university of singapore.
\newblock In \emph{Proceedings of the ACM international conference on image and video retrieval}, pages 1--9, 2009.

\bibitem[Dao et~al.(2023)Dao, Huynh, Zhao, Phung, and Cai]{dao2023open}
Son~D Dao, Dat Huynh, He Zhao, Dinh Phung, and Jianfei Cai.
\newblock Open-vocabulary multi-label image classification with pretrained vision-language model.
\newblock In \emph{2023 IEEE International Conference on Multimedia and Expo (ICME)}, pages 2135--2140. IEEE, 2023.

\bibitem[Deng et~al.(2014)Deng, Luo, Loy, and Tang]{deng2014pedestrian}
Yubin Deng, Ping Luo, Chen~Change Loy, and Xiaoou Tang.
\newblock Pedestrian attribute recognition at far distance.
\newblock In \emph{Proceedings of the 22nd ACM international conference on Multimedia}, pages 789--792, 2014.

\bibitem[Distances(2013)]{distances2013lightspeed}
Cuturi M~Sinkhorn Distances.
\newblock Lightspeed computation of optimal transport.
\newblock \emph{Advances in neural information processing systems}, 26:\penalty0 2292--2300, 2013.

\bibitem[Gao et~al.(2024)Gao, Geng, Zhang, Ma, Fang, Zhang, Li, and Qiao]{gao2024clip}
Peng Gao, Shijie Geng, Renrui Zhang, Teli Ma, Rongyao Fang, Yongfeng Zhang, Hongsheng Li, and Yu Qiao.
\newblock Clip-adapter: Better vision-language models with feature adapters.
\newblock \emph{International Journal of Computer Vision}, 132\penalty0 (2):\penalty0 581--595, 2024.

\bibitem[Gong et~al.(2013)Gong, Jia, Leung, Toshev, and Ioffe]{gong2013deep}
Yunchao Gong, Yangqing Jia, Thomas Leung, Alexander Toshev, and Sergey Ioffe.
\newblock Deep convolutional ranking for multilabel image annotation.
\newblock \emph{arXiv preprint arXiv:1312.4894}, 2013.

\bibitem[Guo et~al.(2023)Guo, Zhang, Qiu, Ma, Miao, He, and Cui]{guo2023calip}
Ziyu Guo, Renrui Zhang, Longtian Qiu, Xianzheng Ma, Xupeng Miao, Xuming He, and Bin Cui.
\newblock Calip: Zero-shot enhancement of clip with parameter-free attention.
\newblock In \emph{Proceedings of the AAAI Conference on Artificial Intelligence}, pages 746--754, 2023.

\bibitem[He et~al.(2023)He, Guo, Dai, Qiao, Shu, Ren, and Xia]{he2023open}
Sunan He, Taian Guo, Tao Dai, Ruizhi Qiao, Xiujun Shu, Bo Ren, and Shu-Tao Xia.
\newblock Open-vocabulary multi-label classification via multi-modal knowledge transfer.
\newblock In \emph{Proceedings of the AAAI Conference on Artificial Intelligence}, pages 808--816, 2023.

\bibitem[Hinton(2015)]{hinton2015distilling}
Geoffrey Hinton.
\newblock Distilling the knowledge in a neural network.
\newblock \emph{arXiv preprint arXiv:1503.02531}, 2015.

\bibitem[Hu et~al.(2022)Hu, Shen, Wallis, Allen-Zhu, Li, Wang, Wang, Chen, et~al.]{hu2022lora}
Edward~J Hu, Yelong Shen, Phillip Wallis, Zeyuan Allen-Zhu, Yuanzhi Li, Shean Wang, Lu Wang, Weizhu Chen, et~al.
\newblock Lora: Low-rank adaptation of large language models.
\newblock \emph{ICLR}, 1\penalty0 (2):\penalty0 3, 2022.

\bibitem[Hu et~al.(2023)Hu, Sun, Sclaroff, and Saenko]{hu2023dualcoop++}
Ping Hu, Ximeng Sun, Stan Sclaroff, and Kate Saenko.
\newblock Dualcoop++: Fast and effective adaptation to multi-label recognition with limited annotations.
\newblock \emph{IEEE Transactions on Pattern Analysis and Machine Intelligence}, 2023.

\bibitem[Hua et~al.(2021)Hua, Mou, Jin, and Zhu]{hua2021multiscene}
Yuansheng Hua, Lichao Mou, Pu Jin, and Xiao~Xiang Zhu.
\newblock Multiscene: A large-scale dataset and benchmark for multiscene recognition in single aerial images.
\newblock \emph{IEEE Transactions on Geoscience and Remote Sensing}, 60:\penalty0 1--13, 2021.

\bibitem[Huynh and Elhamifar(2020)]{huynh2020shared}
Dat Huynh and Ehsan Elhamifar.
\newblock A shared multi-attention framework for multi-label zero-shot learning.
\newblock In \emph{Proceedings of the IEEE/CVF conference on computer vision and pattern recognition}, pages 8776--8786, 2020.

\bibitem[Jia et~al.(2021)Jia, Yang, Xia, Chen, Parekh, Pham, Le, Sung, Li, and Duerig]{jia2021scaling}
Chao Jia, Yinfei Yang, Ye Xia, Yi-Ting Chen, Zarana Parekh, Hieu Pham, Quoc Le, Yun-Hsuan Sung, Zhen Li, and Tom Duerig.
\newblock Scaling up visual and vision-language representation learning with noisy text supervision.
\newblock In \emph{International conference on machine learning}, pages 4904--4916. PMLR, 2021.

\bibitem[Khosla et~al.(2020)Khosla, Teterwak, Wang, Sarna, Tian, Isola, Maschinot, Liu, and Krishnan]{khosla2020supervised}
Prannay Khosla, Piotr Teterwak, Chen Wang, Aaron Sarna, Yonglong Tian, Phillip Isola, Aaron Maschinot, Ce Liu, and Dilip Krishnan.
\newblock Supervised contrastive learning.
\newblock \emph{Advances in neural information processing systems}, 33:\penalty0 18661--18673, 2020.

\bibitem[Kuhn(1955)]{kuhn1955hungarian}
Harold~W Kuhn.
\newblock The hungarian method for the assignment problem.
\newblock \emph{Naval research logistics quarterly}, 2\penalty0 (1-2):\penalty0 83--97, 1955.

\bibitem[Lafon et~al.(2024)Lafon, Ramzi, Rambour, Audebert, and Thome]{lafon2024gallop}
Marc Lafon, Elias Ramzi, Cl{\'e}ment Rambour, Nicolas Audebert, and Nicolas Thome.
\newblock Gallop: Learning global and local prompts for vision-language models.
\newblock In \emph{European Conference on Computer Vision}, pages 264--282. Springer, 2024.

\bibitem[Li et~al.(2018)Li, Zhang, Chen, and Huang]{li2018richly}
Dangwei Li, Zhang Zhang, Xiaotang Chen, and Kaiqi Huang.
\newblock A richly annotated pedestrian dataset for person retrieval in real surveillance scenarios.
\newblock \emph{IEEE transactions on image processing}, 28\penalty0 (4):\penalty0 1575--1590, 2018.

\bibitem[Li et~al.(2023{\natexlab{a}})Li, Wang, Liu, Zeng, Lu, Chen, and Zhou]{li2023patchct}
Miaoge Li, Dongsheng Wang, Xinyang Liu, Zequn Zeng, Ruiying Lu, Bo Chen, and Mingyuan Zhou.
\newblock Patchct: Aligning patch set and label set with conditional transport for multi-label image classification.
\newblock In \emph{Proceedings of the IEEE/CVF International Conference on Computer Vision}, pages 15348--15358, 2023{\natexlab{a}}.

\bibitem[Li et~al.(2023{\natexlab{b}})Li, Wang, Duan, and Li]{li2023clip}
Yi Li, Hualiang Wang, Yiqun Duan, and Xiaomeng Li.
\newblock Clip surgery for better explainability with enhancement in open-vocabulary tasks.
\newblock \emph{arXiv preprint arXiv:2304.05653}, 2023{\natexlab{b}}.

\bibitem[Liero et~al.(2018)Liero, Mielke, and Savar{\'e}]{liero2018optimal}
Matthias Liero, Alexander Mielke, and Giuseppe Savar{\'e}.
\newblock Optimal entropy-transport problems and a new hellinger--kantorovich distance between positive measures.
\newblock \emph{Inventiones mathematicae}, 211\penalty0 (3):\penalty0 969--1117, 2018.

\bibitem[Lin et~al.(2014)Lin, Maire, Belongie, Hays, Perona, Ramanan, Doll{\'a}r, and Zitnick]{lin2014microsoft}
Tsung-Yi Lin, Michael Maire, Serge Belongie, James Hays, Pietro Perona, Deva Ramanan, Piotr Doll{\'a}r, and C~Lawrence Zitnick.
\newblock Microsoft coco: Common objects in context.
\newblock In \emph{Computer Vision--ECCV 2014: 13th European Conference, Zurich, Switzerland, September 6-12, 2014, Proceedings, Part V 13}, pages 740--755. Springer, 2014.

\bibitem[Liu et~al.(2024{\natexlab{a}})Liu, Xue, Gan, Wan, Liang, Deng, Escalera, and Lei]{liu2024cfpl}
Ajian Liu, Shuai Xue, Jianwen Gan, Jun Wan, Yanyan Liang, Jiankang Deng, Sergio Escalera, and Zhen Lei.
\newblock Cfpl-fas: Class free prompt learning for generalizable face anti-spoofing.
\newblock In \emph{Proceedings of the IEEE/CVF Conference on Computer Vision and Pattern Recognition}, pages 222--232, 2024{\natexlab{a}}.

\bibitem[Liu et~al.(2024{\natexlab{b}})Liu, Chen, Guan, Zhou, Zhu, Ye, Fu, and Zhou]{liu2024remoteclip}
Fan Liu, Delong Chen, Zhangqingyun Guan, Xiaocong Zhou, Jiale Zhu, Qiaolin Ye, Liyong Fu, and Jun Zhou.
\newblock Remoteclip: A vision language foundation model for remote sensing.
\newblock \emph{IEEE Transactions on Geoscience and Remote Sensing}, 2024{\natexlab{b}}.

\bibitem[Liu et~al.(2024{\natexlab{c}})Liu, Tan, Tan, Wei, Wang, and Zhao]{liu2024forgery}
Huan Liu, Zichang Tan, Chuangchuang Tan, Yunchao Wei, Jingdong Wang, and Yao Zhao.
\newblock Forgery-aware adaptive transformer for generalizable synthetic image detection.
\newblock In \emph{Proceedings of the IEEE/CVF Conference on Computer Vision and Pattern Recognition}, pages 10770--10780, 2024{\natexlab{c}}.

\bibitem[Liu et~al.(2021)Liu, Zhang, Yang, Su, and Zhu]{liu2021query2label}
Shilong Liu, Lei Zhang, Xiao Yang, Hang Su, and Jun Zhu.
\newblock Query2label: A simple transformer way to multi-label classification.
\newblock \emph{arXiv preprint arXiv:2107.10834}, 2021.

\bibitem[Liu et~al.()Liu, Wen, Liu, Fang, Li, Xu, and Zhang]{liulanguage}
Yicheng Liu, Jie Wen, Chengliang Liu, Xiaozhao Fang, Zuoyong Li, Yong Xu, and Zheng Zhang.
\newblock Language-driven cross-modal classifier for zero-shot multi-label image recognition.
\newblock In \emph{Forty-first International Conference on Machine Learning}.

\bibitem[Loshchilov(2017)]{loshchilov2017decoupled}
I Loshchilov.
\newblock Decoupled weight decay regularization.
\newblock \emph{arXiv preprint arXiv:1711.05101}, 2017.

\bibitem[Luo et~al.(2022)Luo, Ji, Zhong, Chen, Lei, Duan, and Li]{luo2022clip4clip}
Huaishao Luo, Lei Ji, Ming Zhong, Yang Chen, Wen Lei, Nan Duan, and Tianrui Li.
\newblock Clip4clip: An empirical study of clip for end to end video clip retrieval and captioning.
\newblock \emph{Neurocomputing}, 508:\penalty0 293--304, 2022.

\bibitem[Ma et~al.(2024)Ma, Xie, Wang, Fu, Sun, and Zhao]{ma2024text}
Leilei Ma, Hongxing Xie, Lei Wang, Yanping Fu, Dengdi Sun, and Haifeng Zhao.
\newblock Text-region matching for multi-label image recognition with missing labels.
\newblock In \emph{Proceedings of the 32nd ACM International Conference on Multimedia}, pages 6133--6142, 2024.

\bibitem[Ma et~al.(2023)Ma, He, Ran, and Lu]{ma2023transferable}
Peirong Ma, Zhiquan He, Wu Ran, and Hong Lu.
\newblock A transferable generative framework for multi-label zero-shot learning.
\newblock \emph{IEEE Transactions on Circuits and Systems for Video Technology}, 2023.

\bibitem[Mikolov et~al.(2013)Mikolov, Sutskever, Chen, Corrado, and Dean]{mikolov2013distributed}
Tomas Mikolov, Ilya Sutskever, Kai Chen, Greg~S Corrado, and Jeff Dean.
\newblock Distributed representations of words and phrases and their compositionality.
\newblock \emph{Advances in neural information processing systems}, 26, 2013.

\bibitem[Monge(1781)]{monge1781histoire}
G Monge.
\newblock Histoire de l’acad{\'e}mie royale des sciences de paris.
\newblock \emph{De l’Imprimerie Royale}, 1781.

\bibitem[Narayan et~al.(2021)Narayan, Gupta, Khan, Khan, Shao, and Shah]{narayan2021discriminative}
Sanath Narayan, Akshita Gupta, Salman Khan, Fahad~Shahbaz Khan, Ling Shao, and Mubarak Shah.
\newblock Discriminative region-based multi-label zero-shot learning.
\newblock In \emph{Proceedings of the IEEE/CVF international conference on computer vision}, pages 8731--8740, 2021.

\bibitem[Oord et~al.(2018)Oord, Li, and Vinyals]{oord2018representation}
Aaron van~den Oord, Yazhe Li, and Oriol Vinyals.
\newblock Representation learning with contrastive predictive coding.
\newblock \emph{arXiv preprint arXiv:1807.03748}, 2018.

\bibitem[Pennington et~al.(2014)Pennington, Socher, and Manning]{pennington2014glove}
Jeffrey Pennington, Richard Socher, and Christopher~D Manning.
\newblock Glove: Global vectors for word representation.
\newblock In \emph{Proceedings of the 2014 conference on empirical methods in natural language processing (EMNLP)}, pages 1532--1543, 2014.

\bibitem[Qi et~al.(2020)Qi, Zhu, Wang, Zhang, Peng, Wu, Chen, Zhao, Zang, and Mathiopoulos]{qi2020mlrsnet}
Xiaoman Qi, Panpan Zhu, Yuebin Wang, Liqiang Zhang, Junhuan Peng, Mengfan Wu, Jialong Chen, Xudong Zhao, Ning Zang, and P~Takis Mathiopoulos.
\newblock Mlrsnet: A multi-label high spatial resolution remote sensing dataset for semantic scene understanding.
\newblock \emph{ISPRS Journal of Photogrammetry and Remote Sensing}, 169:\penalty0 337--350, 2020.

\bibitem[Radford et~al.(2021)Radford, Kim, Hallacy, Ramesh, Goh, Agarwal, Sastry, Askell, Mishkin, Clark, et~al.]{radford2021learning}
Alec Radford, Jong~Wook Kim, Chris Hallacy, Aditya Ramesh, Gabriel Goh, Sandhini Agarwal, Girish Sastry, Amanda Askell, Pamela Mishkin, Jack Clark, et~al.
\newblock Learning transferable visual models from natural language supervision.
\newblock In \emph{International conference on machine learning}, pages 8748--8763. PMLR, 2021.

\bibitem[Rahman and Khan(2019)]{rahman2019deep}
Shafin Rahman and Salman Khan.
\newblock Deep multiple instance learning for zero-shot image tagging.
\newblock In \emph{Computer Vision--ACCV 2018: 14th Asian Conference on Computer Vision, Perth, Australia, December 2--6, 2018, Revised Selected Papers, Part I 14}, pages 530--546. Springer, 2019.

\bibitem[Rahman et~al.(2019)Rahman, Khan, and Barnes]{rahman2019deep0tag}
Shafin Rahman, Salman Khan, and Nick Barnes.
\newblock Deep0tag: Deep multiple instance learning for zero-shot image tagging.
\newblock \emph{IEEE Transactions on Multimedia}, 22\penalty0 (1):\penalty0 242--255, 2019.

\bibitem[Ren et~al.(2017)Ren, Jin, Lin, Fang, and Yuille]{ren2017multiple}
Zhou Ren, Hailin Jin, Zhe Lin, Chen Fang, and Alan~L Yuille.
\newblock Multiple instance visual-semantic embedding.
\newblock In \emph{BMVC}, 2017.

\bibitem[Ridnik et~al.(2021)Ridnik, Ben-Baruch, Zamir, Noy, Friedman, Protter, and Zelnik-Manor]{ridnik2021asymmetric}
Tal Ridnik, Emanuel Ben-Baruch, Nadav Zamir, Asaf Noy, Itamar Friedman, Matan Protter, and Lihi Zelnik-Manor.
\newblock Asymmetric loss for multi-label classification.
\newblock In \emph{Proceedings of the IEEE/CVF international conference on computer vision}, pages 82--91, 2021.

\bibitem[Ridnik et~al.(2023)Ridnik, Sharir, Ben-Cohen, Ben-Baruch, and Noy]{ridnik2023ml}
Tal Ridnik, Gilad Sharir, Avi Ben-Cohen, Emanuel Ben-Baruch, and Asaf Noy.
\newblock Ml-decoder: Scalable and versatile classification head.
\newblock In \emph{Proceedings of the IEEE/CVF winter conference on applications of computer vision}, pages 32--41, 2023.

\bibitem[Shao et~al.(2024)Shao, Tian, Zhao, and Su]{shao2024explore}
Tong Shao, Zhuotao Tian, Hang Zhao, and Jingyong Su.
\newblock Explore the potential of clip for training-free open vocabulary semantic segmentation.
\newblock In \emph{European Conference on Computer Vision}, pages 139--156. Springer, 2024.

\bibitem[Sinkhorn and Knopp(1967)]{sinkhorn1967concerning}
Richard Sinkhorn and Paul Knopp.
\newblock Concerning nonnegative matrices and doubly stochastic matrices.
\newblock \emph{Pacific Journal of Mathematics}, 21\penalty0 (2):\penalty0 343--348, 1967.

\bibitem[Sun et~al.(2022)Sun, Hu, and Saenko]{sun2022dualcoop}
Ximeng Sun, Ping Hu, and Kate Saenko.
\newblock Dualcoop: Fast adaptation to multi-label recognition with limited annotations.
\newblock \emph{Advances in Neural Information Processing Systems}, 35:\penalty0 30569--30582, 2022.

\bibitem[Sung et~al.(2022{\natexlab{a}})Sung, Cho, and Bansal]{sung2022lst}
Yi-Lin Sung, Jaemin Cho, and Mohit Bansal.
\newblock Lst: Ladder side-tuning for parameter and memory efficient transfer learning.
\newblock \emph{Advances in Neural Information Processing Systems}, 35:\penalty0 12991--13005, 2022{\natexlab{a}}.

\bibitem[Sung et~al.(2022{\natexlab{b}})Sung, Cho, and Bansal]{sung2022vl}
Yi-Lin Sung, Jaemin Cho, and Mohit Bansal.
\newblock Vl-adapter: Parameter-efficient transfer learning for vision-and-language tasks.
\newblock In \emph{Proceedings of the IEEE/CVF conference on computer vision and pattern recognition}, pages 5227--5237, 2022{\natexlab{b}}.

\bibitem[Tan et~al.(2024)Tan, Li, Zhou, Wan, Lei, and Zhang]{tan2024compound}
Hao Tan, Jun Li, Yizhuang Zhou, Jun Wan, Zhen Lei, and Xiangyu Zhang.
\newblock Compound text-guided prompt tuning via image-adaptive cues.
\newblock In \emph{Proceedings of the AAAI Conference on Artificial Intelligence}, pages 5061--5069, 2024.

\bibitem[Xu and Gould(2024)]{xu2024temporally}
Ming Xu and Stephen Gould.
\newblock Temporally consistent unbalanced optimal transport for unsupervised action segmentation.
\newblock In \emph{Proceedings of the IEEE/CVF Conference on Computer Vision and Pattern Recognition}, pages 14618--14627, 2024.

\bibitem[Xu et~al.(2022)Xu, Li, Hsiao, Ho, and Qi]{xu2022open}
S Xu, Y Li, J Hsiao, C Ho, and Z Qi.
\newblock Open vocabulary multi-label classification with dual-modal decoder on aligned visual-textual features.
\newblock \emph{arXiv preprint arXiv:2208.09562}, 2022.

\bibitem[Zhang et~al.(2024)Zhang, Yuan, and Huang]{zhang2024diverse}
Kaixin Zhang, Zhixiang Yuan, and Tao Huang.
\newblock Diverse and tailored image generation for zero-shot multi-label classification.
\newblock \emph{Knowledge-Based Systems}, page 112077, 2024.

\bibitem[Zhang et~al.(2016)Zhang, Gong, and Shah]{zhang2016fast}
Yang Zhang, Boqing Gong, and Mubarak Shah.
\newblock Fast zero-shot image tagging.
\newblock In \emph{2016 IEEE Conference on Computer Vision and Pattern Recognition (CVPR)}, pages 5985--5994. IEEE, 2016.

\bibitem[Zhao et~al.(2021)Zhao, Rao, Wang, Lu, and Zhou]{zhao2021towards}
Wenliang Zhao, Yongming Rao, Ziyi Wang, Jiwen Lu, and Jie Zhou.
\newblock Towards interpretable deep metric learning with structural matching.
\newblock In \emph{Proceedings of the IEEE/CVF International Conference on Computer Vision}, pages 9887--9896, 2021.

\bibitem[Zhong et~al.(2020)Zhong, Zheng, Kang, Li, and Yang]{zhong2020random}
Zhun Zhong, Liang Zheng, Guoliang Kang, Shaozi Li, and Yi Yang.
\newblock Random erasing data augmentation.
\newblock In \emph{Proceedings of the AAAI Conference on Artificial Intelligence}, pages 13001--13008, 2020.

\bibitem[Zhou et~al.(2022{\natexlab{a}})Zhou, Yang, Loy, and Liu]{zhou2022conditional}
Kaiyang Zhou, Jingkang Yang, Chen~Change Loy, and Ziwei Liu.
\newblock Conditional prompt learning for vision-language models.
\newblock In \emph{Proceedings of the IEEE/CVF conference on computer vision and pattern recognition}, pages 16816--16825, 2022{\natexlab{a}}.

\bibitem[Zhou et~al.(2022{\natexlab{b}})Zhou, Yang, Loy, and Liu]{zhou2022learning}
Kaiyang Zhou, Jingkang Yang, Chen~Change Loy, and Ziwei Liu.
\newblock Learning to prompt for vision-language models.
\newblock \emph{International Journal of Computer Vision}, 130\penalty0 (9):\penalty0 2337--2348, 2022{\natexlab{b}}.

\bibitem[Zhou et~al.(2023)Zhou, Pang, Tian, He, and Chen]{zhou2023anomalyclip}
Qihang Zhou, Guansong Pang, Yu Tian, Shibo He, and Jiming Chen.
\newblock Anomalyclip: Object-agnostic prompt learning for zero-shot anomaly detection.
\newblock \emph{arXiv preprint arXiv:2310.18961}, 2023.

\bibitem[Zhu et~al.(2024{\natexlab{a}})Zhu, Liu, Tang, Ge, Liu, Liu, and Cao]{zhu2024query}
Xuelin Zhu, Jian Liu, Dongqi Tang, Jiawei Ge, Weijia Liu, Bo Liu, and Jiuxin Cao.
\newblock Query-based knowledge sharing for open-vocabulary multi-label classification.
\newblock \emph{arXiv preprint arXiv:2401.01181}, 2024{\natexlab{a}}.

\bibitem[Zhu et~al.(2024{\natexlab{b}})Zhu, Ji, Zhao, Wu, and Wang]{zhu2024awt}
Yuhan Zhu, Yuyang Ji, Zhiyu Zhao, Gangshan Wu, and Limin Wang.
\newblock Awt: Transferring vision-language models via augmentation, weighting, and transportation.
\newblock \emph{arXiv preprint arXiv:2407.04603}, 2024{\natexlab{b}}.

\end{thebibliography}
